\documentclass[final,5p,times,authoryear]{elsarticle}

\usepackage{amssymb}
\usepackage[cmex10]{amsmath}
\usepackage{multicol}
\usepackage{multirow}
\usepackage[]{tikz}
\usepackage{tikz-3dplot}
\usetikzlibrary{calc, arrows.meta, positioning, backgrounds, decorations.pathreplacing, quotes, shapes, bayesnet}
\tikzset{>=latex}
\usepackage{array}
\usepackage[colorlinks]{hyperref}
\usepackage[]{subcaption}
\usepackage{booktabs}
\usepackage{xcolor}
\usepackage{soul}

\newcommand\clearrow{\global\let\rowmac\relax}
\clearrow

\journal{Expert Systems with Applications}

\begin{document}


\begin{frontmatter}



\title{One Homography is All You Need: IMM-based Joint Homography and Multiple Object State Estimation}


\author[up]{Paul Johannes Claasen\corref{cor1}}
\ead{pj@benjamin.ng.org.za}

\author[up]{Johan Pieter de Villiers}
\ead{pieter.devilliers@up.ac.za}

\affiliation[up]{organization={Department of Electrical, Electronic and Computer Engineering, University of Pretoria},
            addressline={Lynnwood Road, Hatfield}, 
            city={Pretoria},
            postcode={0028}, 
            country={South Africa}}

\cortext[cor1]{Corresponding author.}

\begin{abstract}
A novel online MOT algorithm, IMM Joint Homography State Estimation (IMM-JHSE), is proposed. IMM-JHSE uses an initial homography estimate as the only additional 3D information, whereas other 3D MOT methods use regular 3D measurements. By jointly modelling the homography matrix and its dynamics as part of track state vectors, IMM-JHSE removes the explicit influence of camera motion compensation techniques on predicted track position states, which was prevalent in previous approaches. Expanding upon this, static and dynamic camera motion models are combined using an IMM filter. A simple bounding box motion model is used to predict bounding box positions to incorporate image plane information. In addition to applying an IMM to camera motion, a non-standard IMM approach is applied where bounding-box-based BIoU scores are mixed with ground-plane-based Mahalanobis distances in an IMM-like fashion to perform association only, making IMM-JHSE robust to motion away from the ground plane. Finally, IMM-JHSE makes use of dynamic process and measurement noise estimation techniques. IMM-JHSE improves upon related techniques, including UCMCTrack, OC-SORT, C-BIoU and ByteTrack on the DanceTrack and KITTI-car datasets, increasing HOTA by 2.64 and 2.11, respectively, while offering competitive performance on the MOT17, MOT20 and KITTI-pedestrian datasets. Using publicly available detections, IMM-JHSE outperforms almost all other 2D MOT methods and is outperformed only by 3D MOT methods---some of which are offline---on the KITTI-car dataset. Compared to tracking-by-attention methods, IMM-JHSE shows remarkably similar performance on the DanceTrack dataset and outperforms them on the MOT17 dataset. The code is publicly available: \url{https://github.com/Paulkie99/imm-jhse}.
\end{abstract}



\begin{keyword}
    multi-object tracking \sep tracking by detection \sep camera calibration \sep camera motion compensation \sep homography estimation


\end{keyword}

\end{frontmatter}


\section{Introduction}
    \noindent The increasing reliability of object detectors \citep{yolo, yolo9000, yolov3, yolov4, yolov9, Ge2021YOLOX:2021, end_end_object} has encouraged research on the topic of tracking objects in the image space, i.e. where objects are tracked in pixel coordinates. Most recent methods have focused on using bounding box information with or without appearance embeddings \citep{Bewley2016SimpleTracking, Zhang2022ByteTrack:Box, botsort, Cao2022Observation-CentricTracking, Yang2022HardSpace}. All of these methods predict the bounding box dimensions of a track in the next frame and associate measured bounding boxes with existing tracks based on these predictions. Yet, it is possible in some cases to obtain an estimate of the location of a bounding box within the ground plane of the imaged 3D space \citep{bhitk}. The ground plane is a coordinate system representing a 2D approximation of the earth's surface in front of the camera, as viewed from above. Leveraging such information may improve association performance. Occlusion and camera motion effects may be mitigated if the target dynamics within the ground plane are reliably estimated and the homographic projection of the ground plane to the image plane is decoupled from the target's motion. This is illustrated in Figure~\ref{fig:illustration}, which depicts two tracks that undergo heavy occlusion in the image plane at time $t$. Although the red track is occluded by the blue track, their locations on the ground plane are still easily separable. Furthermore, camera motion does not influence target association, i.e. additional camera motion compensation is not required if an accurate projection matrix (homography) can be determined at any time step. 

    Despite this, tracks may momentarily leave the ground plane in certain situations, for example, dancers jumping in the DanceTrack dataset \citep{dance}. During these moments, the homographic projection of a target's position in the image plane to the ground plane no longer accurately represents its actual ground-plane location. A multiple object tracking (MOT) algorithm that relies on homographic projections should treat these situations with care. This paper hypothesises that such an MOT algorithm could be made robust to these situations by leveraging a combination of ground- and image-plane filters. Furthermore, it is hypothesised that dynamic process noise estimation can improve performance by helping to account for the divergence of target motion from the prescribed motion model. Similarly, dynamic measurement noise estimation could improve performance by exploiting variance in measurement precision for different targets (i.e. the variability in measurements over time could differ for different targets).

    Few 2D MOT methods experiment with incorporating limited 3D information. One exception is UCMCTrack \citep{ucmc}, which uses an initial homography estimate to perform target tracking in the ground plane. Despite this novelty, UCMCTrack uses the same camera motion compensation technique that has been proven beneficial in state-of-the-art 2D MOT methods like BoT-SORT \citep{botsort}. However, as will be shown, this method does not allow for a dynamic estimate of the homography matrix, induces a forced coupling between camera motion and target state, and precludes joint target and homography state estimation. In contrast, the proposed method decouples camera motion from target state and allows for the joint estimation of target and camera dynamics.

    Many 3D MOT methods use regular 3D measurements \citep{Jiang2024ATracking, Osep2018CombinedScenes, Zhang2024LEGO:Clouds, Wu2022CasA:Clouds, Nagy2024RobMOT:PointCloud, Huang2024BiTrack:Data, Reich2021MonocularTracks, Weng20203DMetrics, Wu20223DAssociation, Wang2024YouTracking, Wang2023TowardsMOT, Wu2023VirtualDetection, Wang2024MCTrack:Driving, Ninh2024CollabMOTTracking, Gao2021GlowEnvironments}, assume the internal camera parameters are known \citep{Choi2013ACamera, Wojek2013MonocularScenes} or attempt to estimate 3D coordinates based on 2D images \citep{Wu2023HybridDevices}. Since IMM-JHSE requires a single initial homography estimate (which only provides a mapping to the 2D world plane), it is considered distinct from these approaches.

    \begin{figure*}[t!]
        \centering
\tdplotsetmaincoords{-60}{-35}

\begin{tikzpicture}
  [
    tdplot_main_coords,
    >=Stealth,
    my dashed/.style={dashed, thin, -},
    my box/.style={thin, gray!70},
    my blue/.style={blue, line cap=round, -{Triangle[width=3*#1]}, line width=#1, shorten >=#1*1.75pt, every node/.append style={fill, circle, inner sep=0pt, minimum size=#1*3.5pt, anchor=center, outer sep=0pt}},
    my label/.append style={midway, font=\scriptsize},
    my vectors/.style={green!50!black, {Stealth[scale=.75]}-{Stealth[scale=.75]}},
    my red/.style={thick, red, line cap=round},
    my grey/.style={gray!70},
    description/.style={draw=gray!100, thick, line cap=round, every node/.style={align=center, font=\scriptsize\sffamily, anchor=north}},
    red grass circle/.style={circle, draw=red!60, fill=red!5, thick, minimum size=8mm, inner sep=1mm},
    red image box/.style={rectangle, draw=red!60, fill=red!5, thick, minimum size=8mm, inner sep=1mm},
    blue grass circle/.style={circle, draw=blue!60, fill=blue!5, thick, minimum size=8mm, inner sep=1mm},
    blue image box/.style={rectangle, draw=blue!60, fill=blue!5, thick, minimum size=8mm, inner sep=1mm},
  ]
  
  \begin{scope}[shift={(0,0)},rotate=-0]
      \coordinate (o) at (0,0,0);
      \path [draw=gray!150, text=gray, fill=green!50, opacity=0.8, text opacity=1] (-4,12,0) coordinate (ag) -- ++(-2,-6,0) coordinate (bg) -- ++(12,0,0) coordinate (cg) -- ++(-0,6,0) coordinate (dg) -- cycle node [pos=.95, above, sloped, anchor=north west] {ground plane}; 
      
      \filldraw[black] (0,0) circle (2pt) node[right] {Optical centre};
      \path [draw=gray!70, text=gray, fill=gray!20, opacity=0.8, text opacity=1] (-3,4,1.25) coordinate (ai) -- ++(0,0,-3.5) coordinate (bi) -- ++(6,0,0) coordinate (ci) -- ++(0,0,3.5) coordinate (di) -- cycle node [pos=.95, above, sloped, anchor=south west] {image plane}; 
      
      \draw[my dashed] (ag)++(1,-2,0) node[red grass circle](rcirc1){\footnotesize $t-1$} -- (o) node[midway, red image box]{\footnotesize $t-1$};
      \draw[my dashed] (ag)++(5,-1.75,0) node[red grass circle](rcirc2){\footnotesize $t$} -- (o) node[midway, red image box]{\footnotesize $t$};
      \draw[dashed] (ag)++(8,-4,0) node[red grass circle](rcirc3){\footnotesize $t+1$} -- (o) node[midway, red image box]{\footnotesize $t+1$};
    
      \draw[my dashed] (ag)++(2,-4,0) node[blue grass circle](bcirc3){\footnotesize $t+1$} -- (o) node[midway, blue image box]{\footnotesize $t+1$};
      \draw[my dashed] (ag)++(5,-2.75,0) node[blue grass circle](bcirc2){\footnotesize $t$} -- (o) node[midway, blue image box]{\footnotesize $t$};
      \draw[dashed] (ag)++(9,-1,0) node[blue grass circle](bcirc1){\footnotesize $t-1$} -- (o) node[midway, blue image box]{\footnotesize $t-1$};
    
      \path [description, dashed, ->, red] (rcirc1) [out=0, in=180] to (rcirc2);
      \path [description, dashed, ->, red] (rcirc2) [out=15, in=180] to (rcirc3);
    
      \path [description, dashed, ->, blue] (bcirc1) [out=-135, in=45] to (bcirc2);
      \path [description, dashed, ->, blue] (bcirc2) [out=-150, in=30] to (bcirc3);

    \end{scope}
        
\end{tikzpicture}
        \caption{Illustration of the potential benefit of using ground plane location estimates. While the blue track occludes the red track in the image plane at time $t$, their ground plane locations are still easily separable.}
        \label{fig:illustration}
    \end{figure*}


    This paper expands upon previous MOT methods. In particular, it makes the following contributions:
        \begin{itemize}
            \item Modelling limitations of a previous method \citep{ucmc} are addressed; see Section~\ref{sec:method}. Briefly, target motion is decoupled from the homography matrix state.
            \item The homography matrix is included in the target state vector. Thus, it is estimated jointly with the target position and velocity.
            \item A static camera motion model (which assumes that the homography matrix does not change over time) is combined with a dynamic camera motion model (which accounts for camera-motion-induced changes in the homography) with an interacting multiple model (IMM) filter.
            \item In addition to applying the IMM to camera motion as mentioned in the point directly above, the proposed method dynamically mixes ground-plane- and image-plane-based association scores, i.e. Mahalanobis distance and buffered intersection over union (BIoU), respectively, for association only.
            \item Dynamic measurement noise estimation is used to estimate the noise associated with a particular track's measured bounding box, and dynamic process noise estimation is used to estimate the noise associated with each camera motion model.
            \item The proposed method outperforms related methods on the DanceTrack \citep{dance} and KITTI-car \citep{kitti} datasets while offering competitive performance on the MOT17 \citep{mot17}, MOT20 \citep{mot20}, and KITTI-pedestrian \citep{kitti} datasets.
        \end{itemize}

    In making the above-mentioned contributions, association performance may be improved in situations where an initial homography or camera calibration estimate is available in lieu of regular 3D measurements and the required hardware. Thus, the affordability of MOT may be improved in some use cases, for example, in pedestrian and car tracking, such as with the KITTI dataset \citep{kitti}, which is important for applications like self-driving cars, or for athletic and gymnastic applications like sports player tracking, dancing or gymnastic performances (e.g. the DanceTrack dataset \citep{dance}). IMM-JHSE may be combined with homography estimation techniques, such as in \citep{bhitk}, to provide regular homography measurements, potentially further improving MOT performance while requiring only a camera-equipped device.
    
    Furthermore, these contributions serve to afford robustness in situations where targets leave the ground plane; enabling tracking algorithms the best of both worlds: exploiting ground plane information when it is reliable, and relying solely on bounding boxes in the image plane otherwise, which is especially useful in applications like sports player tracking, dancing and gymnastic performances.

    The remainder of the paper is structured as follows. Section~\ref{sec:review} examines previous work on image-based tracking. Section~\ref{sec:back} provides more detail on motion models and camera motion compensation in the literature. Section~\ref{sec:method} gives a detailed overview of the proposed method. Section~\ref{sec:exps} provides the experimental setup and results of tests on validation data. Section~\ref{sec:results} reports the results on the MOT17, MOT20, DanceTrack and KITTI test datasets. Finally, Section~\ref{sec:conclusion} concludes the findings of this paper and suggests topics of focus for future work.

\section{Related Work} \label{sec:review}
    \noindent This section provides an overview of previous approaches to the MOT problem to provide context for the contributions presented in this paper. Before reviewing the most relevant literature in Section~\ref{imagebased} and Section~\ref{homotracking}, an overview of 3D MOT methods is provided in Section~\ref{3dmot} to differentiate 3D MOT methods from the proposed solution. This is followed by a brief overview of tracking-by-attention methods in Section~\ref{tba}.

    \subsection{3D MOT}\label{3dmot}
        IMM-JHSE uses only a single initial homography estimate as in \citep{ucmc}. It is important to note that the homography transformation only provides a plane-to-plane mapping. This differentiates it from fully-3D MOT methods which use regular 3D measurements obtained with LiDAR sensors (point clouds) \citep{Jiang2024ATracking, Osep2018CombinedScenes, Zhang2024LEGO:Clouds, Wu2022CasA:Clouds, Nagy2024RobMOT:PointCloud, Huang2024BiTrack:Data}, RGB-D cameras \citep{Jafari2014Real-timeCameras, Luber2010LearningData}, 3D detections \citep{Reich2021MonocularTracks, Weng20203DMetrics, Wu20223DAssociation, Wang2024YouTracking, Wang2023TowardsMOT, Wu2023VirtualDetection, Wang2024MCTrack:Driving}, odometry \citep{Gao2021GlowEnvironments} and stereo or binocular cameras \citep{Ninh2024CollabMOTTracking, Cao2015RobustCamera, Ess2009RobustPlatform}.

        In addition, the methods in \citep{Choi2013ACamera, Wojek2013MonocularScenes} use a Reversible Jump Markov Chain Monte Carlo particle filter with various feature representations, including skin colour, depth images, facial detection \citep{Choi2013ACamera} and explicit occlusion reasoning \citep{Wojek2013MonocularScenes}. Furthermore, these methods consider different camera motion models than the one proposed here (which explicitly considers camera position and rotation) and assume that internal camera parameters are known. Finally, \citep{Wojek2013MonocularScenes} further makes use of odometry information.

        Furthermore, the method in \citep{Wu2023HybridDevices} attempts to perform 3D tracking without any of these measurements. However, the authors take an iterative approach to camera motion estimation. In contrast, IMM-JHSE integrates camera motion and its dynamics into the state vector to estimate it jointly with track states. They also rely on detecting vanishing points and lines parallel to the horizon. In addition, the multi-mode filter they employ only considers the target state and needs to be reinitialised with previous measurements when a mode switch occurs. Finally, multiple hypothesis tracking (MHT) performs association, whereas IMM-JHSE uses a linear assignment algorithm.

        It could be argued that homography information is akin to 3D measurements. However, it should be emphasised that only one initial homography estimate is required for IMM-JHSE as opposed to the regular measurements exploited by 3D MOT methods. In addition, the homography is only a plane-to-plane mapping. As such, the authors of this paper still consider IMM-JHSE a 2D MOT method since two 2D planes are considered.

    \subsection{Tracking by attention}\label{tba}
        \noindent Instead of the pre-defined anchor boxes used in several popular object detectors, Carion et al. \citep{end_end_object} decode object positions from a set of learned position embeddings using a transformer decoder. Meinhardt et al. \citep{trackformer} extend this approach to MOT by using previous detections as queries in subsequent frames, resulting in the Trackformer architecture. Whereas Trackformer works on a frame-by-frame basis, Global Tracking Transformers, as proposed by Zhou et al. \citep{gtr}, encode detections in a sliding temporal window of frames. A transformer decoder then associates new detections with these encodings. Having access to an entire measurement sequence, the deep data association module proposed by Pinto et al. \citep{transformer_dda} eschews the transformer decoder and relies solely on an encoder to associate detections within the given sequence with one another. End-to-end attention-based methods have recently succeeded on MOT leaderboards \citep{fusiontrack, Gao2024MeMOTR:Tracking, Gao2024MultiplePrediction}. However, these methods are computationally expensive, especially considering they often train their own detectors in addition to training the attention-based mechanism in an end-to-end approach.

    \subsection{Image-based Tracking}\label{imagebased}
    
        \noindent Tracking the motion of objects from a video has been of interest as early as 1998 when Isard and Blake \citep{condensation} developed the condensation algorithm for tracking objects represented by a curve or sets of curves with learned motion models. Their method propagates conditional density distributions similar to the approach of particle filtering. However, tracking multiple objects simultaneously with their method may become infeasible, especially for complex multi-modal posterior densities where large samples or particles may be required to obtain accurate estimates. Besides, multi-object tracking requires solving the association problem -- deciding which observations belong to which tracked object -- which becomes increasingly difficult when the tracked objects are near one another, partially or fully occluding each other. Multiple object tracking (MOT) methods in the literature focus more on the association problem than on detection or individual target tracking. 
        
        Given a set of object detections, Li et al. \citep{Zhang2008GlobalFlows} model the association problem with a cost-flow network, with the solution obtained by a minimum-cost flow algorithm. They also explicitly model occluded observations, which is achieved by simply comparing differences in detection locations and scale to predetermined thresholds. While the approach is reported to be real-time and yields favourable results on the  Context-Aware Vision using Image-based Active Recognition (CAVIAR) \citep{caviar} dataset, object motion is not explicitly modelled. As a result, the method may perform poorly when object motion is highly dynamic. Berclaz et al. \citep{Berclaz2011MultipleOptimization} similarly model object trajectories with a flow model consisting of a discretised 2D spatial grid (representing the image area) at each time step. They use the K-shortest path algorithm to solve their linear programming formulation of the association problem. Motion is modelled by only allowing an occupied node at the previous time step to transition to a specific neighbourhood around that node in the current time step. As with \citep{Zhang2008GlobalFlows}, this is an unsatisfactory solution in situations where objects are highly dynamic. 
        
        With the increasing reliability and popularity of several object detectors \citep{yolo, yolo9000, yolov3, yolov4, yolov9, Ge2021YOLOX:2021, end_end_object}, modern methods have primarily made use of bounding box information. Bewley et al. \citep{Bewley2016SimpleTracking} emphasise the importance of detection quality and use the Hungarian algorithm to perform data association based on the intersection-over-union (IoU) of detected bounding boxes with those predicted by Kalman-filtered tracks (specifically, predicted bounding boxes are constructed from the predicted bounding box centre x- and y-coordinates, height and aspect ratio). Their method has been influential in promoting the use of Kalman filters to estimate object motion distributions and/or the Hungarian algorithm (or other linear assignment algorithms) to perform association: these have become the status quo in \citep{modelling_ambiguous, invisible, Wang2020TowardsTracking, Zhang2021FairMOT:Tracking, giao, Zhang2022ByteTrack:Box, botsort}. Taking the minimalist approach of the original SORT algorithm one step further, Bergmann et al. \citep{bells} rely purely on the regression head of their detector network (or ``Tracktor'') to perform tracking. However, they show that performance improves with camera motion compensation and appearance-feature-based re-identification, suggesting that explicit motion models (and appearance information) can still be beneficial. Supporting this idea, Khurana et al. \citep{invisible} use motion models that consider monocular depth estimates to maintain occluded tracks. SORT is improved upon in \citep{Wojke2017SimpleMetric} by integrating appearance information, resulting in DeepSORT. They train a convolutional neural network (CNN) to discriminate between pedestrians and keep each track's normalised appearance embeddings of the last 100 frames. Track association is performed by finding the minimum cosine distance of the current detected appearance embedding with the history of embeddings for each of the previous tracks. Additionally, they use motion information by obtaining the Mahalanobis distance between detections and predicted Kalman filter states, consisting of bounding box centre positions, aspect ratios, heights and their corresponding velocities. In SiMIlarity LEarning for Occlusion-Aware Multiple Object Tracking (SMILEtrack), Hsiang et al. \citep{Wang2022SMILEtrack:Tracking} make use of a Siamese-based network which incorporates elements inspired by vision transformers (specifically, self-attention computed on image patches) to extract appearance features from detected bounding boxes. However, it does not achieve a convincing improvement in higher-order tracking accuracy (HOTA) \citep{Luiten2021HOTA:Tracking} compared to ByteTrack, despite its superior computational speed. Leveraging multiple-camera traffic surveillance systems, another method explicitly models inter-vehicle occlusion using reconstructed-cuboid projections instead of relying on similarity learning \citep{vehicle_occlusion}. It is noted that the use of camera motion compensation has become ubiquitous, with applications in \citep{bells, invisible, modelling_ambiguous, giao, Zhang2022ByteTrack:Box, botsort, Wang2022SMILEtrack:Tracking, interframe}.  
        
        Whereas previous methods separately perform detection and appearance embedding extraction, Wang et al. \citep{Wang2020TowardsTracking} design a model which performs both of these simultaneously. They report that similar results can be achieved compared to state-of-the-art SDE (separate detection and embedding) methods at reduced computational cost. Zhang et al. \citep{Zhang2021FairMOT:Tracking} show that such multi-task networks must be designed carefully since the tasks require different features and feature dimensions and that the anchors used for object detection can introduce ambiguity in the re-identification features. They introduce a carefully designed anchor-free model to simultaneously perform object detection and appearance embedding extraction, which outperforms the state-of-the-art methods in tracking metrics and frame rate.
        
        Shifting the focus from appearance embeddings, Zhang et al. \citep{Zhang2022ByteTrack:Box} note that it is beneficial to perform association in two steps in their method called ByteTrack: the first step takes only high-confidence detections into account, while the remaining unmatched detections, as well as the lower-confidence detections, are associated in a second association step. They note that using the IoU is essential in the second association step since other features may become unreliable for low-confidence detections, which may be occluded and/or blurred. Still, re-identification or other features may be used in the first step. By altering the Kalman filter state vector, Aharon et al. \citep{botsort} improve upon ByteTrack. Instead of tracking the aspect ratio of bounding boxes, they perform better by directly tracking bounding box widths and heights. Similar to \citep{giao, Zhang2021FairMOT:Tracking, Wang2020TowardsTracking}, an exponential moving average mechanism is used to update the appearance states of tracklets -- the features of which are obtained by a re-identification network from the FastReID library \citep{He2020FastReID:Re-identification}. Specifically, the bags of tricks \citep{Luo2019BagRe-Identification} stronger baseline with a ResNeSt50 \citep{Zhang2020ResNeSt:Networks} backbone is used. Accordingly, they name their method BoT-SORT. Although tracking width and height may improve performance, Cao et al. \citep{Cao2022Observation-CentricTracking} note that previous motion models are estimation-centric, allowing for rapid error accumulation over time, especially when no observations are available, and motion is non-linear. This implies that previous approaches are more sensitive to the noise in state estimation than the noise in measurements. They propose observation-centric SORT (OC-SORT), which incorporates a re-update step when an object is re-identified, during which the Kalman filter is updated with virtual detections generated based on the last-known and currently detected bounding box state. In addition, they use a novel association criterion that considers the change in motion direction that each new detection would induce, which is termed observation-centric momentum (OCM). Finally, Yang et al. \citep{Yang2022HardSpace} buffers object bounding boxes to expand the matching space of subsequent detections in their Cascaded Buffered IoU (C-BIoU) tracker. A cascaded matching approach is used such that small buffers are used in the first association step, and larger buffers are used in the second step. Despite its simplicity, Their method performs superior to DeepSORT, SORT, ByteTrack and OC-SORT.

    \subsection{Tracking Involving Homography}\label{homotracking}
    
        \noindent Various methods have used homography between multiple camera views to increase the robustness of the tracking of sports players. Since a particular player may be occluded in one camera view but not in another, using multiple cameras increases the robustness of a player tracking system to player occlusions. Examples of methods which make use of inter-camera homographies may be found in \citep{Iwase2003TrackingViews, Iwase2004ParallelImages, Eshel2008HomographyCrowd, Seo2009HumanHomography, Sainan2018AthletesStudy, Khan2006AConstraint}. However, few approaches consider the homography between a single camera view and the playing field. Notably, Hayet et al. \citep{Hayet2005ATracking} incrementally update the homography between the camera view and an imaged soccer field and explicitly use it to track player locations on the playing field with a Kalman filter. Recently, Maglo et al. \citep{Maglo2023IndividualView} compute the homography to display player positions, but the tracking is performed without the aid of the homography. Instead, they rely heavily on appearance features extracted by a model fine-tuned offline after tracklet generation (on the corresponding bounding boxes). As a result, their method is not suitable for online operation. Most recently, Yi et al. \citep{ucmc} manually annotated the homographic projections between the image and ground planes of various datasets. They then perform tracking and association in the ground plane. However, as will be shown in Section~\ref{sec:method}, camera motion explicitly influences target position in their method -- which this paper regards as a modelling error.

\section{Background: Camera Motion Compensation and Motion Models}\label{sec:back}
        \noindent The majority of modern approaches to MOT model target motion in the image plane \citep{Bewley2016SimpleTracking, Zhang2022ByteTrack:Box, botsort, Cao2022Observation-CentricTracking, Yang2022HardSpace}. Particularly, the near-constant velocity model is usually employed:
            \begin{equation}
                \label{image_dyn}
                \mathbf{x}^I_t=\mathbf{x}^I_{t-1}+\Dot{\mathbf{x}}^I_{t-1}\Delta t,
            \end{equation}
        where $\mathbf{x}^I_t=\begin{bmatrix}
            x^{I}_t & y^{I}_t & w^{I}_t & h^{I}_t
        \end{bmatrix}^\top$ represents the centre x- and y-coordinate, width and height (in pixels) of the target's bounding box at time $t$, respectively. The superscript $I$ indicates that the state vector $\mathbf{x}^I_t$ is represented in the image plane. The vector encompassing the velocity of each state component is denoted by $\Dot{\mathbf{x}}$, and $\Delta t$ represents the sampling period --- in this case, it is equal to the reciprocal of the video frame rate. The additive noise term has been omitted for brevity. 
        
        The current state is conditioned on the previous state, with measurements in the form of detected bounding boxes directly related to it. This model is graphically represented in Figure~\ref{fig:graphical_model_I},
        \begin{figure}[hbt!]
            \centering
                \begin{tikzpicture}
                    \node[circle,draw=black,inner sep=5pt] (x1) at (0,0) {\normalsize $\mathbf{x}^I_0$};
                    \node[circle,draw=black,fill=green!10,inner sep=5pt] (obs1) at (0,-1.5) {\normalsize $\mathbf{y}^I_0$};
                    
                    \node[text width=0.6cm] (dots) at (1.5,0) {$\cdots$};
                    
                    \node[circle,draw=black,inner sep=5pt] (x2) at (3,0) {\normalsize $\mathbf{x}^I_t$};
                    \node[circle,draw=black,fill=green!10,inner sep=5pt] (obs2) at (3,-1.5) {\normalsize $\mathbf{y}^I_t$};
                    
                    \path [draw,->] (x1) edge [left] node [right] {} (dots);
                    \path [draw,->] (dots) edge [left] node [right] {} (x2);
                    \path [draw,->] (x1) edge [below] (obs1);
                    \path [draw,->] (x2) edge [below] (obs2);
            \end{tikzpicture}
            \caption{The graphical model usually employed in modern MOT approaches. $\mathbf{x}^I_t$ represents the state of a bounding box element (x/y position, width or height), and $\mathbf{y}^I_t$ denotes the corresponding measurement, i.e. an element of the detected bounding box.}
            \label{fig:graphical_model_I}
        \end{figure}
        and assumes that the camera remains stationary. As a result, several approaches \citep{bells, invisible, modelling_ambiguous, giao, Zhang2022ByteTrack:Box, botsort, Wang2022SMILEtrack:Tracking, interframe} make use of camera motion compensation after prediction with (\ref{image_dyn}). Specifically, a matrix 
            \begin{equation}
                \label{affine_mat}
                \mathbf{\hat{A}}_t = \left[\begin{array}{c|c} 
                            \mathbf{R}^A_t\in \mathcal{R}^{2\times 2} & \mathbf{t}_t\in\mathcal{R}^{2\times 1} \\
                            \mathbf{0}^{1\times 2} & 1
                            \end{array}\right]
            \end{equation}
        is estimated such that
            \begin{equation}
                \label{motion_comp_xy}
                \begin{bmatrix}
                    x^{I'}_t \\
                    y^{I'}_t \\
                    1
                \end{bmatrix} = \mathbf{\hat{A}}_t\begin{bmatrix}
                    x^{I}_t \\
                    y^{I}_t \\
                    1
                \end{bmatrix}
            \end{equation}
        and
            \begin{equation}
                \label{motion_comp_wh}
                \begin{bmatrix}
                    w^{I'}_t \\
                    h^{I'}_t \\
                \end{bmatrix} = \mathbf{R}^A_t\begin{bmatrix}
                    w^{I}_t \\
                    h^{I}_t \\
                \end{bmatrix},
            \end{equation}
        where the prime superscript denotes the camera-motion-compensated state component. $\mathbf{\hat{A}}_t$ is estimated with the camera calibration module of the OpenCV \citep{opencv} library as in \citep{botsort}. Most previous approaches make use of optical flow features and the RANSAC \citep{ransac} algorithm, although \citep{ucmc} makes use of the enhanced correlation coefficient (ECC) optimisation method.

        UCMCTrack \citep{ucmc} instead models target motion in the ground plane:
            \begin{equation}
                \label{ground_dyn}
                \mathbf{x}^W_t=\mathbf{x}^W_{t-1}+\Dot{\mathbf{x}}^W_{t-1}\Delta t + \mathbf{w}^{W},
            \end{equation}
        where $\mathbf{x}^W_t=\begin{bmatrix}
            x^W_t & y^W_t
        \end{bmatrix}^\top$ represents the target's x- and y-coordinates, respectively. The superscript $W$ indicates that the state is represented in the ground (or world) plane. The additive noise term $\mathbf{w}^W$ is sampled from a zero-mean multivariate normal distribution with covariance $\mathbf{Q}^{W,\mathbf{x}}$ (i.e. the process noise covariance matrix) determined by\citep{ucmc}:
                \begin{equation}
                    \mathbf{Q}^{W,\mathbf{x}}=\mathbf{\Gamma}\operatorname{diag}\left(\sigma_x,\sigma_y\right)\mathbf{\Gamma}^T,
                \end{equation}
            where 
            \begin{equation}
                \mathbf{\Gamma}=\left[\begin{array}{ll}
                \frac{\Delta t^2}{2} & 0 \\
                \Delta t & 0 \\
                0 & \frac{\Delta t^2}{2} \\
                0 & \Delta t
                \end{array}\right],
            \end{equation}
            and $\sigma_x$ and $\sigma_y$ are process noise compensation factors along the x- and y-axes, respectively. The superscript $\mathbf{x}$ in $\mathbf{Q}^{W,\mathbf{x}}$ differentiates this noise term from those associated with the homography components in (\ref{homog_dyn}), (\ref{homog_static}). 
            
            UCMCTrack manually annotates the homographic projection relating the ground plane to the starting frame of each video on which they evaluate their method. Let $\mathbf{H}^L$ represent this initial projection, then 
            \begin{equation}
                \label{projection}
                \begin{bmatrix}
                    x^{I}_t \\
                    x^{I}_t + 0.5 \times h^{I}_t \\
                    1
                \end{bmatrix} = \operatorname{norm}\left(\mathbf{H}^L\begin{bmatrix}
                                        x^{W}_t \\
                                        y^{W}_t \\
                                        1
                                    \end{bmatrix}\right),
            \end{equation}
        where $\operatorname{norm}(\begin{bmatrix}
            x & y & z 
        \end{bmatrix}^\top) = \begin{bmatrix}
            x/z & y/z & 1
        \end{bmatrix}^\top$. For details regarding the theory of the pinhole camera model employed here, the reader is directed to \citep{bhitk, multiple_view_geometry}. The left-hand side of (\ref{projection}) will subsequently be referred to as the bottom-centre bounding box coordinates. As in \citep{ucmc}, these coordinates are selected to represent the projection of an object's ground plane position since it is where the object is expected to touch the ground (e.g. if the object is a person, this coordinate represents where their feet touch the ground).
        
        UCMCTrack does not consider how the projection matrix $\mathbf{H}^L$ evolves over time. Because of this, existing tracks' positions must be corrected to compensate for camera motion. In particular, UCMCTrack makes use of a post-prediction camera motion compensation step. First, the bottom-centre bounding box location is obtained by applying (\ref{projection}) to the predicted ground plane coordinates. The approximate centre x- and y-coordinates of the bounding box in the image plane are then obtained by subtracting half of the last-known height from the y-coordinate of this projection. This estimate is then used in (\ref{motion_comp_xy}) to approximate the x- and y-coordinates of the target bounding box in the current frame after camera motion compensation (CMC) has been applied. CMC is also applied to the width and height of the last-known bounding box (i.e. the bounding box which has been associated with the particular track most recently) as in (\ref{motion_comp_wh}). Finally, the inverse of (\ref{projection}) is used to obtain the camera-motion-compensated ground plane coordinates. Ignoring the effect of bounding box width and height, this update step can roughly but concisely be described by
            \begin{equation}
                \label{ucmc_motioncomp}\mathbf{x}^{W'}_t=\operatorname{norm}\left(\left(\mathbf{H}^L\right)^{-1}\mathbf{\hat{A}}_t\operatorname{norm}\left(\mathbf{H}^L\mathbf{x}^W_t\right)\right).
            \end{equation}
        One might interpret (\ref{ucmc_motioncomp}) to imply that the target position in the ground plane is in some way dependent on camera motion, which is counter-intuitive to the knowledge that a target's movement does not depend on camera motion. Target and camera motion models are treated independently in the proposed solution.
        
        \section{The Proposed Solution}\label{sec:method}
        \noindent Let $\mathbf{H}^W_t$ represent the projection matrix as it evolves over time. Its evolution may be described by \citep{bhitk}
            \begin{equation}
                \label{homog_dyn}
                \mathbf{H}^W_t = \mathbf{\mathbf{\hat{A}}}_t\mathbf{H}^W_{t-1} + \mathbf{\Tilde{Q}}_t^{W,\mathbf{H}},
            \end{equation}
        with $\mathbf{H}^W_0=\mathbf{H}^L$, and $\mathbf{\Tilde{Q}}_t^{W,\mathbf{H}}$ is an additive noise term which is estimated as described in Section~\ref{noise_est}. The superscript $\mathbf{H}$ differentiates this noise term from that of the ground plane position components, and the tilde differentiates it from the static motion model introduced in (\ref{homog_static}). Equation~\ref{homog_dyn} describes how the homographic projection changes as a function of the previous projection and the estimated camera motion. Importantly, this is independent of and does not influence target motion. Note that the derivation of (\ref{homog_dyn}) in \citep{bhitk} assumes that ground plane points are static. While this is not the case where the target is a dynamic object, the motion model is applied nonetheless, especially since the process noise estimation procedure in Section~\ref{noise_est} dynamically adjusts the expected error statistics.
        
        A graphical model which combines Equation~\ref{ground_dyn} and Equation~\ref{homog_dyn} with the measurement model in (\ref{projection}) is shown in Figure~\ref{fig:graphical_model_G}. Here, $\mathbf{x}^I_t$ represents the bottom-centre bounding box position of the target detection. According to Bayesian network theory, the target and camera projection dynamics are independent while the target bounding box is unobserved, but they become dependent when conditioned on $\mathbf{x}_t^I$ \citep{BIS}. Thus, the temporal evolution of the ground plane and homography states are independent but are related through a common measurement in the image plane. This model may allow for a more accurate estimation of a target's ground plane motion since its ground plane position prediction is not directly influenced by camera motion --- only the position of its projection in the image plane.

        \begin{figure}[hbt!]
            \centering
            \begin{tikzpicture}
                \node[circle,draw=black,inner sep=5pt] (X1) at (0,1.5) {\normalsize$\mathbf{x}^W_0$};
                \node[circle,draw=black,inner sep=5pt] (H1) at (0,-1.5) {\normalsize$\mathbf{H}^W_0$};
                \node[circle,draw=black,fill=green!10,inner sep=5pt] (obs1) at (0,0) {\normalsize$\mathbf{x}^I_0$};
                
                \node[text width=0.6cm] (dots_X) at (2,1.5) {$\cdots$};
                \node[text width=0.6cm] (dots_H) at (2,-1.5) {$\cdots$};
                
                \node[circle,draw=black,inner sep=5pt] (X2) at (4,1.5) {\normalsize$\mathbf{x}^W_t$};
                \node[circle,draw=black,inner sep=5pt] (H2) at (4,-1.5) {\normalsize$\mathbf{H}^W_t$};
                \node[circle,draw=black,fill=green!10,inner sep=5pt] (obs2) at (4,0) {\normalsize$\mathbf{x}^I_t$};
                
                \path [draw,->] (X1) edge [left] node [right] {} (dots_X);
                \path [draw,->] (dots_X) edge [left] node [right] {} (X2);
                
                \path [draw,->] (H1) edge [left] node [right] {} (dots_H);
                \path [draw,->] (dots_H) edge [left] node [right] {} (H2);
                
                \path [draw,->] (X1) edge [below] (obs1);
                \path [draw,->] (H1) edge [below] (obs1);
                
                \path [draw,->] (X2) edge [below] (obs2);
                \path [draw,->] (H2) edge [below] (obs2);
                
            \end{tikzpicture}
            \caption{The graphical model representing ground plane and camera motion as independent stochastic processes that become dependent when conditioned on the observed bounding box measurement.}
            \label{fig:graphical_model_G}
        \end{figure}

        The authors of \citep{ucmc} note that in scenes with relatively low camera motion, their method performs better without the camera motion compensation step in (\ref{ucmc_motioncomp}). This may be because the estimated affine matrix (\ref{affine_mat}) is inaccurate, and noise is not considered in (\ref{ucmc_motioncomp}). To remedy this, the method proposed in this paper includes an additive noise term in (\ref{homog_dyn}) (see Section~\ref{statedef} and Section~\ref{noise_est}). Additionally, to better model the evolution of $\mathbf{H}^W_t$ when there is no camera motion, the following motion model is included:
            \begin{equation}
                \label{homog_static}
                \mathbf{H}^W_t = \mathbf{H}^W_{t-1} + \mathbf{\bar{Q}_t^{W,\mathbf{H}}},
            \end{equation}
        where $\mathbf{\bar{Q}_t^{W,\mathbf{H}}}$ is an additive noise term, and the bar differentiates it from that in (\ref{homog_dyn}). For each track, two filters are run in parallel. One of these makes use of (\ref{homog_dyn}), and the other makes use of (\ref{homog_static}). The states of these filters are fused in an IMM filter as described in \ref{immappendix}.
        
        Furthermore, it has been shown that using buffered intersection over union (BIoU) during association can increase performance in scenes where the camera or the tracks are highly dynamic \citep{Yang2022HardSpace}. Besides, there may be times when the tracks leave the ground plane (for example, jumps or flips in the DanceTrack dataset \citep{dance}), causing their measured ground plane positions to fluctuate wildly. During these times, it may be better to perform association with an image-based association measure such as the BIoU. Even if tracks do not leave the ground plane, they may be occluded, which could cause partial detections that manifest similar behaviour to when tracks leave the ground plane. Nevertheless, considering bounding box dimensions during association increases performance, as shown in Section~\ref{sec:results}. Therefore, the proposed method uses the simple bounding box motion model proposed in \citep{Yang2022HardSpace}. To perform association, the ground plane and image-based association scores are taken into account proportionally according to the predicted model likelihood \citep{imm}:
            \begin{equation}
                \label{imm_proba}
                \mu_{t \mid t-1}^{i}=P\left(m_t^{i} \mid \mathbf{x}^I_{1: t-1}\right)=\sum_{j=1}^r p_{ji} \mu_{t-1}^{j},
            \end{equation}
        where $\mu_{t-1}^{j}$ denotes the probability that the target motion matches model $m^{j}$ at time $t-1$, $p_{ji}$ denotes the probability of a transition from model $m^{j}$ to model $m^{i}$, $r$ denotes the number of models under consideration and $\mathbf{x}^I_{1: t-1}$ represents all past observations. This is explained further in Section~\ref{assoc_alg}.

        Finally, the proposed method uses dynamic measurement and process noise estimation, significantly improving performance (see Section~\ref{noise_est}).

        \subsection{State Definition and Motion Models}\label{statedef}
            \noindent The ground plane state of a track includes its position and velocity along both ground plane axes. Furthermore, this paper proposes including the elements of the projection matrix that relate the track's ground plane position to its position in the image plane in the state vector. Thus, the ground plane state is defined as 
            \begin{equation}
                \label{ground_state}
                \mathbf{x}_t = \begin{bmatrix}
                    x^{W}_t\\ \Dot{x}^{W}_t \\ y^{W}_t \\ \Dot{y}^{W}_t \\ \mathbf{h}^{W}_1 \\\mathbf{h}^{W}_2 \\ \mathbf{h}^{W}_3
                \end{bmatrix},
            \end{equation}
            where $\mathbf{h}^{W}_1$, $\mathbf{h}^{W}_2$ and $\mathbf{h}^{W}_3$ are the first, second and third columns of $\mathbf{H}^W_t$, respectively.

            Note that although each track is treated as having its own unique projection matrix, in reality, there is a single projection matrix that relates the image plane to the ground plane. Examining Figure~\ref{fig:graphical_model_G}, this would imply that the current tracks' ground plane positions become dependent upon receiving an image plane measurement. However, this work handles each track independently to simplify track management. 

            The proposed method uses the constant velocity model for the ground plane dimensions as in (\ref{ground_dyn}). For the homography state elements, both the dynamics models of (\ref{homog_dyn}) and (\ref{homog_static}) are combined with an IMM filter to remain robust in both static and dynamic camera motion scenarios. The reader is referred to \citep{imm, Blackman1999DesignSystems, imm_genovese, Blom1988TheCoefficients} for details regarding IMM filters, but an overview is provided in \ref{immappendix}. The process noise covariance matrix corresponding to the homography state elements is determined as Section~\ref{noise_est} explains.

            When a track's position is predicted without a measurement update, it is said to be \textit{coasting}, and it is in the \textit{coasted} state. Otherwise, it is \textit{confirmed}. In the image plane, the bounding box prediction is independent of the state vector in (\ref{ground_state}) while the track is \textit{confirmed}. This paper makes use of the simple motion model presented in \citep{Yang2022HardSpace}, which averages bounding box velocities over a buffer of previous measurements and adds this average to the previous bounding box measurement:
                \begin{equation}\label{bbox_pred}
                    \mathbf{x}^M_{t\mid t-1}=\mathbf{x}^M_{t-1} + \frac{1}{n - 1}\sum_{i=t-n+1}^{t-1}\mathbf{x}^M_{i}-\mathbf{x}^M_{i-1},
                \end{equation}
            $n$ denotes the size of the buffer, which stores the bounding box measurements $\mathbf{x}^M_{t-n:t-1}$ previously associated with the particular track. The superscript $M$ indicates a measurement, except in $\mathbf{x}^M_{t\mid t-1}$ which refers to the predicted measurement at time $t$ given the measurements from $t-n$ to $t-1$. Each measurement is stored as $\begin{bmatrix}
                l^{I} & t^{I}& r^{I}& b^{I}
            \end{bmatrix}$, where the components denote the left x-, top y-, right x- and bottom y-coordinate of the corresponding bounding box, respectively. The maximum size of the bounding box measurement buffer is fixed to $n=5$. When fewer than two measurements are in the buffer, the last-known bounding box is used as the prediction.

\begin{figure*}[!t]
                    \centering
                    \subfloat[Track 3 starts \textit{coasting}.]{\includegraphics[width=0.45\textwidth]{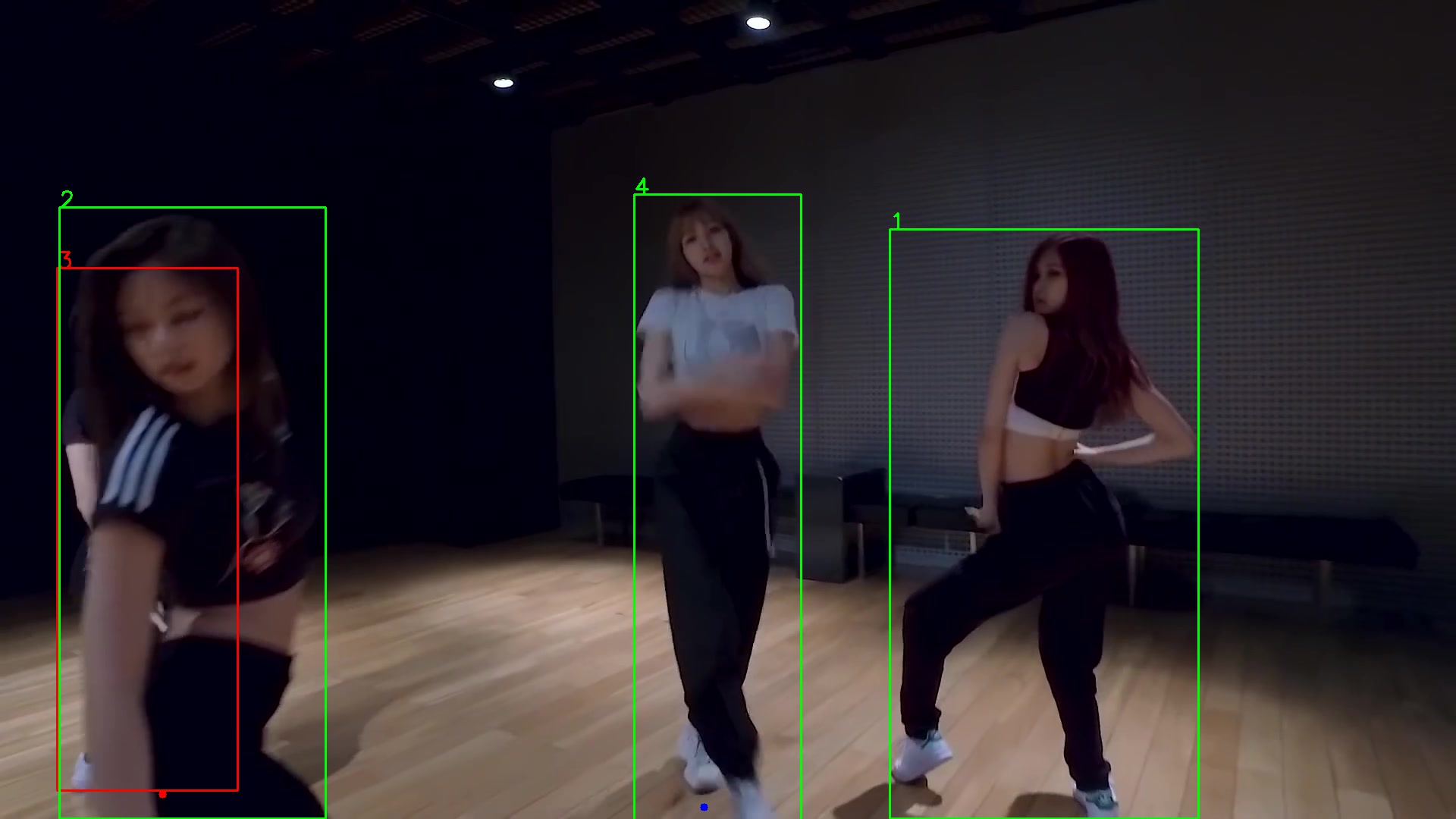}%
                    \label{fig:coupled_pred_a}}
                    \hfil
                    \subfloat[The bounding box prediction for track 3 is coupled to its predicted ground plane position.]{\includegraphics[width=0.45\textwidth]{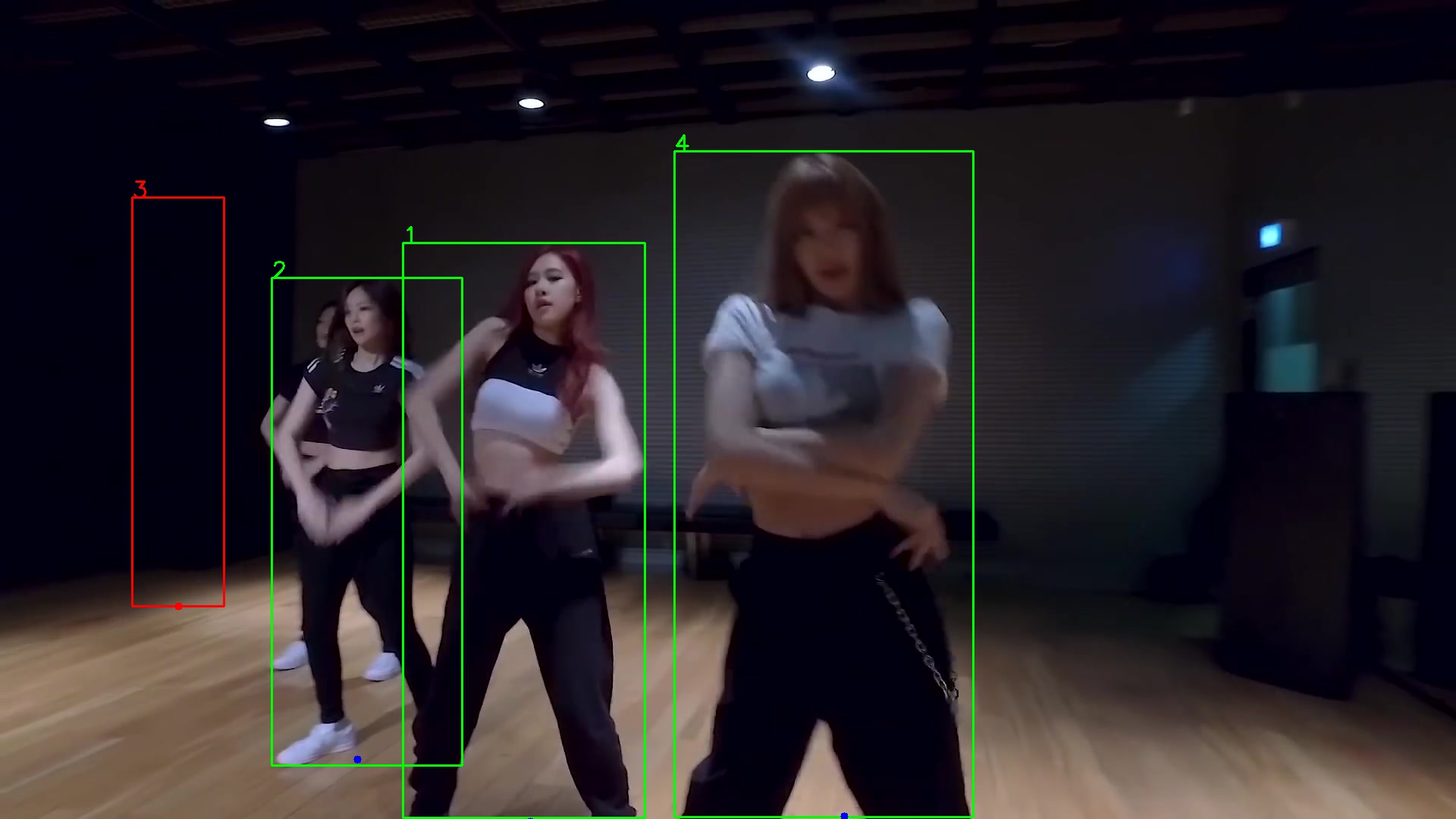}%
                    \label{fig:coupled_pred_b}}
                    \hfil
                    \subfloat[Track 3 is re-identified.]{\includegraphics[width=0.45\textwidth]{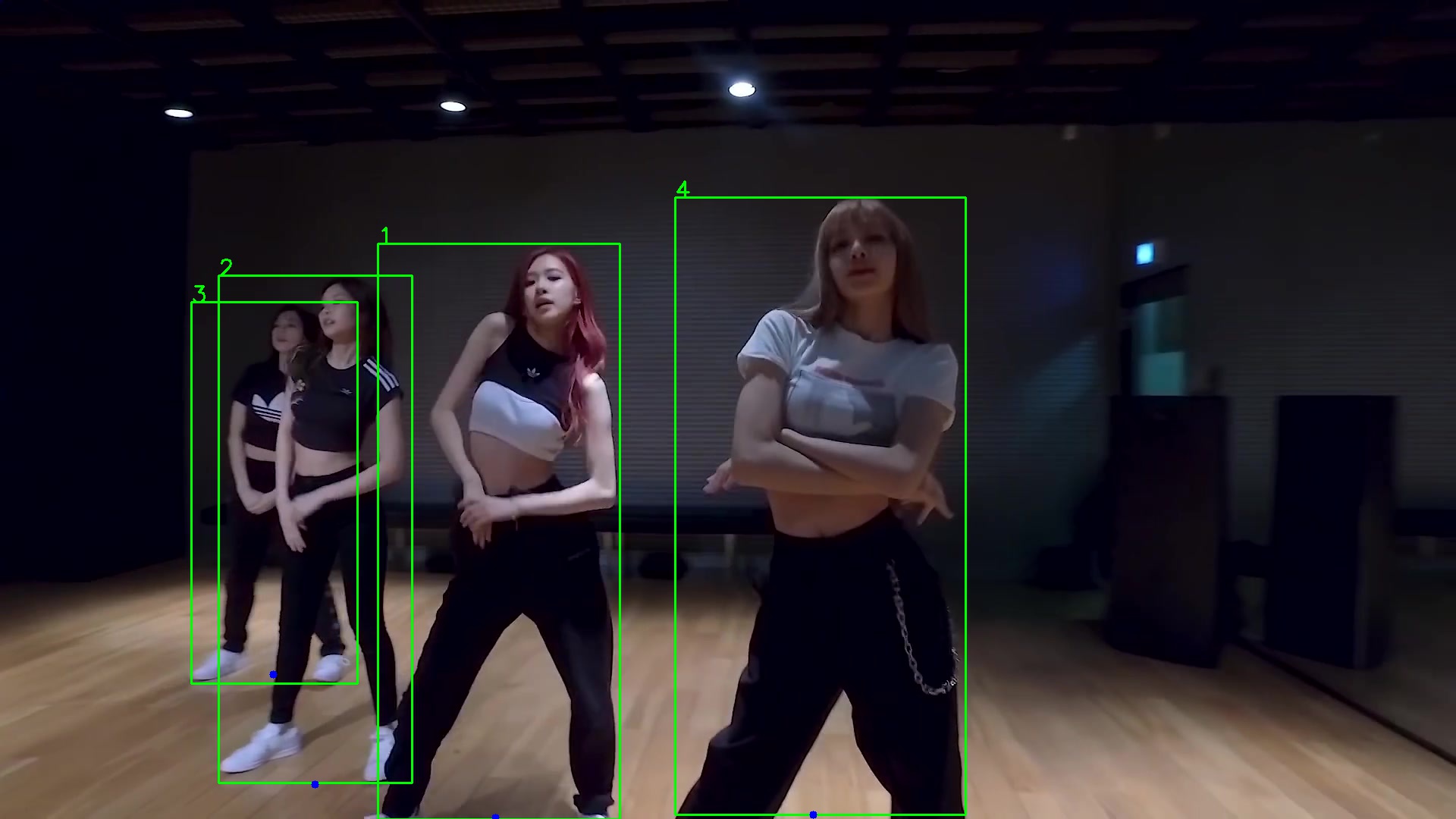}%
                    \label{fig:coupled_pred_c}}
                    \caption{Showing how coupling bounding box predictions to ground plane state estimates may benefit re-identification. Green bounding boxes represent confirmed tracks, while red represents \textit{coasted} tracks. The track ID is shown at the top-left above each bounding box. Blue dots represent the ground plane position of the corresponding confirmed track as projected into the image plane, while a red dot represents the same for a \textit{coasted} track. By coupling the bounding box prediction for track 3 to its predicted ground plane position, it can be re-identified with the BIoU score. The predicted bounding box goes off-screen without this coupling, and the track cannot be re-identified.}
                    \label{fig:coupled_pred}
                \end{figure*}

            When a track is \textit{coasted}, the predicted bounding box is coupled to the predicted ground plane position. Specifically, the bottom-centre coordinates of the predicted bounding box are set to the projection of the predicted ground plane position in the image plane. The width and height of the predicted bounding box are determined by (\ref{motion_comp_wh}), where the last-confirmed bounding box width and height are propagated through time. This is illustrated in Figure~\ref{fig:coupled_pred}, where a track is successfully re-identified after a period of occlusion due to this technique. When a measurement is associated with a track after a period of being \textit{coasted}, the previous buffer of (\ref{bbox_pred}) is cleared and contains only the newly associated detection.
                
        \subsection{Measurement Model}
            \noindent Let $\bar{x}^{I}$ and $\bar{y}^{I}$ denote the bottom-centre x- and y- bounding box coordinates in the image plane as obtained in (\ref{projection}). A track's ground plane coordinates $x^{W}$ and $y^{W}$ are related to these image coordinates through its homographic projection state elements:
            \begin{equation}
                \label{meas_model}
                \begin{bmatrix}
                    \bar{x}^{I}_t \\
                    \bar{y}^{I}_t \\
                    1
                \end{bmatrix} = \operatorname{norm}\left(\mathbf{H}^W_t\begin{bmatrix}
                                        x^{W}_t \\
                                        y^{W}_t \\
                                        1
                                    \end{bmatrix}\right) + \mathbf{R}_t= h(\mathbf{x}) + \mathbf{R}_t,
            \end{equation}
            $\mathbf{R}_t$ is a noise term which is dynamically estimated as described in Section~\ref{noise_est}, but initially set to
                \begin{equation}\label{initial_r0}
                    \mathbf{R}_0=\left[\begin{array}{cc}
\left(\sigma_m w^{I}_0\right)^2 & 0 \\
0 & \left(\sigma_m h^{I}_0\right)^2
\end{array}\right]
                \end{equation}  
            as in \citep{ucmc}, where $\sigma_m=0.05$.

            Let the vector $\mathbf{b}$ represent the unnormalised projection of the ground plane position:
            \begin{equation}
                \mathbf{b} = \begin{bmatrix}
                                        b_1 \\
                                        b_2 \\
                                        b_3
                                    \end{bmatrix} = \mathbf{H}^W_t\begin{bmatrix}
                                        x^{W}_t \\
                                        y^{W}_t \\
                                        1
                                    \end{bmatrix}
            \end{equation}
            and
            \begin{equation}
                \mathbf{H}^W_t=\begin{bmatrix}
                    h_1 & h_2  & h_3\\
                    h_4 & h_5 & h_6\\
                    h_7 & h_8 & h_9
                \end{bmatrix}.
            \end{equation}
            As shown in \ref{deriv}, the Jacobian matrix of (\ref{meas_model}) with respect to the ground plane coordinates is
            \begin{equation}\label{jac_g}
                \frac{\partial\left(x^{I}_t, y^{I}_t\right)}{\partial\left(x^{W}_t, y^{W}_t\right)} = \gamma\begin{bmatrix}
                    (h_1-h_7x^I_t) & (h_2-h_8x^I_t)\\
                    (h_4-h_7y^I_t) & (h_5-h_8y^I_t)
                \end{bmatrix},
            \end{equation}
        where $\gamma = b_3^{-1}$.

        Furthermore, the Jacobian matrix with respect to the homography matrix elements is:
            \begin{equation}\label{jac_h}
                \begin{split}
                    &\frac{\partial\left(x^{I}_t, y^{I}_t\right)}{\partial\left(h_1,h_4,h_7,h_2,h_5,h_8,h_3,h_6\right)} =\\ &\gamma\begin{bmatrix}
                        x^W_t & 0 & -x^I_tx^W_t & y^W_t & 0 & -x^I_ty^W_t & 1 & 0\\
                        0 & x^W_t & -y^I_tx^W_t & 0 & y^W_t & -y^I_ty^W_t & 0 & 1
                    \end{bmatrix}.
                \end{split}
            \end{equation}
        In practice, $\mathbf{H}^W_t$ is normalised with respect to $h_9$ such that $h_9=1$. Therefore, the partial derivative with respect to $h_9$ is $0$.

        \subsection{Dynamic Noise Estimation}\label{noise_est}
            \noindent Instead of attempting to determine a static process noise covariance matrix for the dynamics models in (\ref{homog_dyn}) and (\ref{homog_static}), the proposed method obtains dynamic estimates of these parameters. This allows IMM-JHSE to exploit the possibility of different targets having different noise parameters, and allows these estimates to be robust against irregularities such as when targets leave the ground plane or occlude one another (often resulting in partial detections). The same is performed for the measurement noise covariance matrix associated with the measurement model (\ref{meas_model}) -- this is shown to increase performance in Table~\ref{tab:dance_ablation}. 
            
            From \citep{adaptive_noise}, consider a generalised representation of the process noise covariance matrix $\mathbf{Q}_t$, which can represent either the static motion model of (\ref{homog_static}), or the dynamic motion model of (\ref{homog_dyn}). This matrix is the expected value of the outer product of the state process noise vector $\mathbf{w}_t$:
                \begin{equation}
                    \mathbf{Q}_t=\mathrm{E}\left[\mathbf{w}_t\mathbf{w}_t^T\right].
                \end{equation}
            Similarly, the measurement noise covariance matrix is the outer product of the state measurement noise vector $\mathbf{v}_t$:
                \begin{equation}
                    \mathbf{R}_t=\mathrm{E}\left[\mathbf{v}_t\mathbf{v}_t^T\right].
                \end{equation}

            When a measurement update is performed, the measurement noise covariance matrix for the current time step is estimated by \citep{adaptive_noise}:
                \begin{equation}\label{meas_noise_current}
                    \mathrm{E}\left[\mathbf{v}_t\mathbf{v}_t^T\right]=\mathbf{\varepsilon}_t \mathbf{\varepsilon}_t^T + \mathbf{J}_t\mathbf{P}_{t|t-1}\mathbf{J}_t^T,
                \end{equation}
            where $\mathbf{J}_t$ is the Jacobian matrix with respect to the predicted state components obtained from (\ref{jac_g}), (\ref{jac_h}). $\mathbf{P}_{t|t-1}$ is the \textit{a priori} state covariance matrix. Let $\mathbf{\bar{x}}^{M}_t$ denote the measured bottom-centre x- and y- bounding box coordinate vector, then $\mathbf{\varepsilon}_t=\left[\mathbf{\bar{x}}^{M}_t-h\left(\mathbf{\hat{x}}_{t|t}\right)\right]$ is the residual between the measurement and the updated state vector estimate $\mathbf{\hat{x}}_{t|t}$ projected into the image plane. This paper makes use of a moving window to average noise covariance estimates such that
                \begin{equation}\label{meas_noise}
                    \mathbf{R}_t=\frac{1}{m}\sum_{i=t-m+1}^t\mathrm{E}\left[\mathbf{v}_i\mathbf{v}_i^T\right],
                \end{equation}
            where $m=5$. Note that the Kalman filter update step requires an estimate of $\mathbf{R}_t$, yet $\mathbf{R}_t$ depends on the residual of the updated state vector in (\ref{meas_noise_current}). To use as much information as possible at the current time step, a ``dummy'' Kalman filter update is first performed to obtain the quantity in (\ref{meas_noise_current}). The actual update step is then performed with $\mathbf{R}_t$ obtained from (\ref{meas_noise}).

            After $\mathbf{R}_t$ is obtained from (\ref{meas_noise}), the resulting Kalman gain $\mathbf{K}_t$ is used to obtain an estimate of the process noise covariance matrix for the current time step \citep{adaptive_noise}:
                \begin{equation}\label{process_noise_current}
                    \mathrm{E}\left[\mathbf{w}_{t} \mathbf{w}_{t}^T\right]=\mathrm{E}\left[\mathbf{K}_t\left(\mathbf{d}_t \mathbf{d}_t^T\right) \mathbf{K}_t^T\right]=\mathbf{K}_t \mathrm{E}\left[\mathbf{d}_t \mathbf{d}_t^T\right] \mathbf{K}_t^T,
                \end{equation}
            where $\mathbf{d}_t=\left[\mathbf{\bar{x}}^{M}_t-h\left(\mathbf{\hat{x}}_{t|t-1}\right)\right]$ is the measurement innovation and $\mathbf{\hat{x}}_{t|t-1}$ is the predicted \textit{a priori} state estimate. Similar to (\ref{meas_noise}), a moving window is used to average process noise covariance estimates obtained from (\ref{process_noise_current}). The process noise covariance matrix for the homography state elements, which will be used in the subsequent prediction at time step $t + 1$, is obtained by
                \begin{equation}\label{homog_dyn_noise}
                    \mathbf{Q}^{W,\mathbf{H}}_t=\frac{1}{m}\sum_{i=t-m+1}^t\mathrm{E}\left[\mathbf{w}_{t} \mathbf{w}_{t}^T\right].
                \end{equation}
            Only the matrix elements which correspond to homography state elements are extracted from (\ref{process_noise_current}), such that the final process covariance matrix for the entire state is $\mathbf{Q}_t=\operatorname{diag}(\mathbf{Q}^{W,\mathbf{x}},\mathbf{Q}^{W,\mathbf{H}}_t)$. $\mathbf{Q}^{W,\mathbf{H}}_t$ refers either to $\mathbf{\Tilde{Q}}^{W,\mathbf{H}}_t$ or to $\mathbf{\bar{Q}}^{W,\mathbf{H}}_t$, depending on the filter motion model in question.

            \subsubsection{Convergence}
                Dynamic noise estimates are obtained independently for each track; it is therefore not sufficient to show empirical evidence of convergence for a small number of select tracks to prove convergence in general, yet an investigation of convergence over all tracks is complicated by the possibility that different targets have different true noise covariance matrices. Furthermore, because target motion may be intermittently irregular, the true noise covariance matrices are not expected to be static. We refer to \citep{adaptive_noise} for the original derivation of the above formulae and for their analysis of filter convergence. Furthermore, under the interpretation that the noise estimates are obtained by a sample covariance estimator, the Cramér-Rao lower bound can be shown to be monotonically decreasing as the window size increases \citep{Kay1993}. Therefore, convergence is theoretically possible with the assumption that the true noise covariance is static.

                The rigorous analysis of the convergence of these estimates is deferred to future work.

        \subsection{Association and Track Management}\label{assocalg}
            \noindent This section describes the association algorithm and how the BIoU and Mahalanobis distances are combined to form robust association scores between tracks and candidate detections. The Hungarian algorithm (specifically, the Jonker-Volgenant variation \citep{lapjv}) uses these scores in cascaded matching stages to obtain track-detection associations. There are three association stages, with association score thresholds $\alpha_1, \alpha_2$ and $\alpha_3$, respectively. The values of the hyperparameters introduced in this section are obtained as described in Section~\ref{sec:exps}.

            \subsubsection{Track Management}
                Before the association algorithm is described, it is necessary to define the track states that determine by which branch of the algorithm a specific track will be considered. When a track is initialised, it is in the \textit{tentative} state. From the \textit{tentative} state, a track becomes \textit{confirmed} if associated with a detection for two consecutive time steps after initialisation; otherwise, it is deleted if it is not associated with any detection for two consecutive time steps. If a \textit{confirmed} track's position is predicted without a measurement update (i.e. it is not associated with any detection), it is \textit{coasting} and is in the \textit{coasted} state. If a measurement is associated with a \textit{coasted} track, its state changes again to \textit{confirmed}. If a \textit{coasted} track is not associated with a measurement within $\Omega$ time steps, it is deleted.

            \subsubsection{The Association Algorithm}\label{assoc_alg}
                Figure~\ref{fig:alg_flow} depicts the association algorithm graphically. At the start of a time step, the detections for the corresponding video frame are obtained, and all existing tracks' positions are predicted by their ground- and image-plane filters. From here, \textit{confirmed} and \textit{coasted} tracks are considered by two cascaded association stages. Before the first stage, detections are separated into high- and low-confidence detections by the confidence thresholds $d_{\text{high}}$ and $d_{\text{low}}$, as introduced in \citep{Zhang2022ByteTrack:Box}. Specifically, detections with confidence greater than or equal to $d_{\text{high}}$ are considered high-confidence, and those with confidence less than $d_{\text{high}}$ but greater than or equal to $d_{\text{low}}$, are considered low-confidence. Detections with confidences below $d_{\text{low}}$ are discarded.

                During the first association stage, high-confidence detections are associated with the \textit{confirmed} and \textit{coasted} tracks, with the association score
                    \begin{equation}
                        \label{stage1_score}
                        P(D)\times \text{BIoU}(\mathbf{x}^M_t, \mathbf{x}^M_{t|t-1})\times d_{\text{conf}},
                    \end{equation}
                where $d_{\text{conf}}$ refers to the detection confidence value output by the detector, and $D$ is the normalised Mahalanobis distance as in \citep{ucmc}:
                    \begin{equation}
                        D=\mathbf{d}_t^T\mathbf{S}_t^{-1}\mathbf{d}_t + \ln |\mathbf{S}_t|,
                    \end{equation}
                where $\mathbf{S}_t=\mathbf{J}_t\mathbf{P}_{t|t-1}\mathbf{J}_t^T + \mathbf{R}_{t-1}$, $|\mathbf{S}_t|$ denotes the determinant of $\mathbf{S}_t$, $\ln$ is the natural logarithm and $\mathbf{R}_{t-1}$ is the measurement noise covariance estimate (\ref{meas_noise}) obtained in the previous update step. For reasons that will be explained subsequently, $D$ is converted to a probability by 
                    \begin{equation}
                        P(D)=1 - \operatorname{CDF}(D),
                    \end{equation}
                where $\operatorname{CDF}$ is the cumulative distribution function of the chi-squared distribution, which is defined to have $24$ degrees of freedom (determined experimentally). As a result, $P(D)$ is in the range $[0, 1]$. Finally, $\text{BIoU}(\mathbf{x}^M_t, \mathbf{x}^M_{t|t-1})$ denotes the BIoU between a given detection $\mathbf{x}^M_t$ and the bounding box predicted by (\ref{bbox_pred}). The BIoU is calculated as in \citep{Yang2022HardSpace}, where the IoU is calculated after the measured and predicted bounding box widths and heights are each scaled by $2b+1$; in this paper, only a single buffer scale parameter, $b$, is used.

                After the first association stage, the remaining unassociated detections are combined with the low-confidence detections. These are associated with the remaining unassociated \textit{confirmed} and \textit{coasted} tracks, with the association score
                    \begin{equation}
                        \label{stage2_score}
                        \left(\mu^{I}_{t|t-1}\text{BIoU}(\mathbf{x}^M_t, \mathbf{x}^M_{t|t-1})+\mu_{t|t-1}^{W}P(D)\right)d_{\text{conf}},
                    \end{equation}
                where $\mu^{I}_{t|t-1}$ and $\mu^{W}_{t|t-1}$ refer to the predicted model probabilities of the image and ground plane filters, respectively. These weighting factors are determined by (\ref{imm_proba}). From (\ref{stage2_score}), it should be clear that $D$ is converted to probability $P(D)$ to allow the Mahalanobis distance to be mixed with the BIoU on the same scale.
                
                This association stage manifests the filter architecture depicted in Figure~\ref{fig:method_graphical}. While the ground plane filters are mixed explicitly with an IMM filter as established in the literature \citep{Blom1988TheCoefficients, imm, Blackman1999DesignSystems, imm_genovese} and explained in \ref{immappendix}, the image and ground plane filters are handled independently. Transition probabilities between these filters are only defined to use the predicted model likelihood (\ref{imm_proba}) during association. Each model probability is initialised as $\mu^I_0=\mu^W_0=0.5$. When a measurement is associated with a track, these model probabilities are updated \citep{imm}:
                    \begin{equation}
                        \mu^{i}_t=\frac{\mu^i_{t|t-1}\Lambda^i_t}{\sum_j\mu^j_{t|t-1}\Lambda^j_t},
                    \end{equation}
                where $\Lambda^i_t$ is the likelihood of the associated measurement given model $i\in\{I,W\}$. Specifically, the proposed method uses $\Lambda^I_t=\text{BIoU}(\mathbf{x}^M_t, \mathbf{x}^M_{t|t-1})$ and $\Lambda^W_t=P(D)$. This is another reason for converting $D$ to a probability: so that the likelihoods $\Lambda^I_t$ and $\Lambda^W_t$ are on the same scale.
                
                The second association stage helps to avoid tracks from \textit{coasting} when their detections in the image plane diverge from their expected value as predicted by the ground plane filter. This may occur due to partial occlusion or irregular, rapid object motion. It may also occur when ground plane filter state estimates are inaccurate, notably when a track is re-identified after a period during which it was \textit{coasted} \citep{Cao2022Observation-CentricTracking}. In these cases, the predicted model likelihood $\mu^W_{t|t-1}$ is expected to be close to zero, while the predicted model likelihood $\mu^I_{t|t-1}$ is expected to be close to one. Thus, the association score (\ref{stage2_score}) will be dominated by the BIoU, and the track may still be associated despite low $P(D)$. Note that \citep{Cao2022Observation-CentricTracking} introduced a re-update step to mitigate the effect of state estimation noise and that the second association stage proposed here could be considered an alternative to their solution.

                The image and ground plane filters are not mixed explicitly because a target's ground plane state estimate should not be significantly influenced when the BIoU dominates the association score since this often implies motion away from the ground plane (e.g. a jump) or a partial detection, both of which are expected to result in an inaccurate measurement of the ground plane position. However, if the primary goal is to reduce the effect of state estimation noise as in \citep{Cao2022Observation-CentricTracking}, mixing ground and image plane state estimates may be beneficial.

                After the second association stage, the remaining unassociated tracks are placed in the \textit{coasted} state. Of the unassociated detections, those with detection confidence greater than $d_{\text{high}}$ are used in the third association stage to be associated with the \textit{tentative} tracks. The third association stage also uses the association score in (\ref{stage2_score}). The unassociated detections that remain after this stage are used to initialise new tracks.  

                The hybrid association strategy outlined above is expected to afford IMM-JHSE enhanced robustness to situations where targets unexpectedly leave the ground plane or undergo irregular motion that would otherwise cause their motion to diverge from the prescribed ground-plane motion model.
                \begin{figure}[hbt!]
                    \centering
                    \resizebox{0.8\columnwidth}{!}{%
                    \tikzset{every loop/.style={min distance=10mm,looseness=1}}
                    \begin{tikzpicture}
                    \tikzstyle{camera_style} = [circle, text centered, draw=black, fill=red!20, inner sep=2pt, text width=6mm]
                    \tikzstyle{biou_style} = [circle, text centered, draw=black, fill=blue!20, inner sep=2pt, text width=6mm]
                    
                    \node (static) [camera_style] {\tiny Static Camera Filter};
                    \node (dynamic) [camera_style, right=of static] {\tiny Dynamic Camera Filter};
                    
                    \draw[->] (static) to[bend left] (dynamic);
                    \path[->] (static) to[bend left] node[midway,above,inner sep=2pt] {\tiny$p_{s,d}$} (dynamic);
                    
                    \draw[->] (static) to[out=-60,in=-120,loop] node[midway,below,inner sep=2pt] {\tiny$p_{s,s}$} ();
                    
                    \draw[->] (dynamic) to[bend left] (static);
                    \path[->] (dynamic) to[bend left] node[midway,below,inner sep=2pt] (pds) {\tiny$p_{d,s}$} (static);
                    
                    \draw[->] (dynamic) to[out=-60,in=-120,loop] node[midway,below,inner sep=2pt] {\tiny$p_{d,d}$} ();
                    
                    \draw[thick,dotted] ($(static.north west)+(-0.25,0.5)$) rectangle ($(dynamic.south east)+(0.25,-1.2)$);
                    \node () at ($(static.north west)+(1.1,0.4)$) {\tiny Ground Plane IMM Filter};
                    
                    \node (left_imm) at ($(static.west)+(0.05,-0.35)$) {};
                    
                    \path let \p1 = (left_imm) in node (immrect_centre) at ($0.5*(static.west)+(-0.125,0)+0.5*(dynamic.east)+(0.125,0)+(0,-1.2)+(0,\y1)$) {};
                    
                    \draw[->, dotted, thick] ($(immrect_centre.north)+(0,2.4)$) to[out=60,in=120,loop] node[midway,above,inner sep=2pt] {\tiny$p_{W,W}$} ();
                    
                    \node (biou) at ($(pds.south)+(-0,-2)$) [biou_style] {\tiny Image Plane Filter};
                    \draw[->,thick,dotted] (immrect_centre) to[bend left] (biou);
                    \path[->] (immrect_centre) to[bend left] node[midway,right,inner sep=2pt] () {\tiny$p_{W,I}$} (biou);
                    \draw[->,thick,dotted] (biou) to[bend left] (immrect_centre);
                    \path[->] (biou) to[bend left] node[midway,left,inner sep=2pt] () {\tiny$p_{I,W}$} (immrect_centre);
                    \draw[->,thick,dotted] (biou) to[out=-60,in=-120,loop] node[midway,below,inner sep=2pt] {\tiny$p_{I,I}$} ();
                    
                    \end{tikzpicture}
                    }
                    \caption{A graphical representation of the proposed method. Dynamic and static camera motion models are combined with an IMM filter. During association, the normalised Mahlanbobis distances from this filter and the corresponding BIoU scores are weighted proportionally as described in Section~\ref{assoc_alg}. Since the states of the image and ground plane filters are not mixed, the state transitions between them are depicted with dashed lines. The transition probabilities are determined as explained in Section~\ref{sec:exps}.}
                    \label{fig:method_graphical}
                 \end{figure}
            \begin{figure*}[hbt!]
                \centering
                \resizebox{2\columnwidth}{!}{%
                \tikzstyle{startstop} = [rectangle, rounded corners, 
                minimum width=3cm, 
                minimum height=1cm,
                text centered, 
                draw=black, 
                fill=red!10, align=center]
                
                \tikzstyle{trackio} = [chamfered rectangle, 
                minimum width=3cm, 
                minimum height=1cm, text centered, 
                draw=black, fill=gray!20, align=center]
                
                \tikzstyle{detio} = [chamfered rectangle, 
                minimum width=3cm, 
                minimum height=1cm, text centered, 
                draw=black, fill=blue!20, align=center]
                
                \tikzstyle{process} = [rectangle, 
                minimum width=3cm, 
                minimum height=1cm, 
                text centered, 
                text width=3cm, 
                draw=black, 
                fill=yellow!20]
                
                \tikzstyle{decision} = [signal, signal to=west and east, minimum width=3cm, minimum height=1cm, text centered, draw=black, fill=cyan!20, align=center]
                
                \tikzstyle{arrow} = [thick,->,>=stealth]
                
                \begin{tikzpicture}[node distance=2cm]
                
                \node (dets1) [detio] {High Confidence\\Detections};
                \node (confcoast) [trackio, below=2cm of dets1] {Confirmed Tracks\\Coasted Tracks};
                
                \path[->] (dets1.south) to node[midway] (leftofstage1) {} (confcoast.north);
                \node (start) [startstop, left=of leftofstage1] {Start Time Step $t$};
                \draw[arrow] (start) |- (dets1.west) node(get_dets)[process, midway]{Get Detections};
                \draw[arrow] (start) |- (confcoast.west) node (pred_tracks) [process, midway]{Predict All Track Positions};
                \node (stage1) [process, right=2cm of leftofstage1] {Association Stage 1\\w. Equation~\ref{stage1_score}};
                \node (stage2) [process, right=3.5cm of stage1] {Association Stage 2\\w. Equation~\ref{stage2_score}};
                \node (dets2) [detio] at (dets1 -| stage2) {Low Confidence\\Detections};
                \draw[arrow] (get_dets.north) |- ($(dets2.north)+(0,0.3)$) --(dets2.north);
                
                \node (stage1input) at ($(leftofstage1.center)+(1.75,0)$) {};
                
                \draw[arrow] (dets1.east) -| (stage1input.center) -- (stage1.west);
                \draw[-] (confcoast.east) -| (stage1input.center);
                
                \draw[arrow] ($(stage1.north east)-(0,0.2)$) -- ($(stage2.north west)-(0,0.2)$) node[midway,above]{Remaining Detections};
                \draw[arrow] ($(stage1.south east)+(0,0.2)$) -- ($(stage2.south west)+(0,0.2)$) node[midway,below]{Remaining Tracks};
                
                \draw[arrow] (dets2.south) -- (stage2.north);
                
                \path[->] (stage1.east) to node[midway] (midstage12) {} (stage2.west);
                \node (meas_update) [process] at (midstage12 |- confcoast) {Measurement Updates};
                \draw[arrow] (stage1.south) |- (meas_update.west) node[midway,below]{Confirmed Tracks};
                \draw[arrow] (stage2.south) |- (meas_update.east) node[midway,below]{Confirmed Tracks};
                
                \draw[-] ($(stage2.north east)-(0,0.2)$) -- ($(stage2.north east)+(3.5,-0.2)$) node[midway,above]{Remaining Tracks} node[] (end_track_s2) {};
                \draw[-] ($(stage2.south east)+(0,0.2)$) -- ($(stage2.south east)+(3.5,0.2)$) node[midway,below]{Remaining Detections} node[] (end_det_s2) {};
                
                \node [decision,above=1cm of end_track_s2] (coastdel) {Track Older Than $\Omega$?};
                \node [decision,below=1cm of end_det_s2] (coastdet) {High Confidence Detection?};
                
                \draw[arrow] (end_track_s2.center) -- (coastdel);
                \draw[arrow] (end_det_s2.center) -- (coastdet);
                
                \draw[arrow] (coastdel.east) -- ($(coastdel.east)+(1,0)$) node[midway, above] () {Yes} node[right] () {Delete};
                \draw[arrow] (coastdel.north) -- ($(coastdel.north)+(0,1)$) node[midway, left] () {No} node[above] () {Mark Coasted};
                
                \draw[arrow] (coastdet.east) -- ($(coastdet.east)+(1,0)$) node[midway, above] () {No} node[right] () {Delete};
                
                \node (tentative) [trackio, below=3cm of confcoast] {Tentative Tracks};
                \draw[arrow] (pred_tracks) |- (tentative);
                \node (stage3) [process] at ($(stage1 |- tentative)+(0,1)$) {Association Stage 3\\w. Equation~\ref{stage2_score}};
                
                \node(stage3trackin)[] at($(stage3.south west)+(-0.4,+0.2)$) {};
                \draw[arrow] (tentative) -| (stage3trackin.center) -- ($(stage3.south west)+(-0,+0.2)$);
                
                \node(stage3detin)[] at($(stage3.north west)+(-0.4,-0.2)$) {};
                
                \draw[arrow] (coastdet.south) |- ($(stage3detin.center)+(0,1.5)$) node[midway,below]{Yes} -- (stage3detin.center) -- ($(stage3.north west)+(0,-0.2)$);
                
                \node [decision,below=2.6cm of stage2] (deltent) {Unassociated for 2 Time Steps?};
                \node [process,below=1cm of deltent] (initnew) {Initiate New Tracks};
                \node [decision,below=0.5cm of initnew] (conftent) {Associated for 2 Time Steps?};
                
                \draw[arrow] (deltent.east) -- ($(deltent.east)+(1,0)$) node[midway, above] () {Yes} node[right] () {Delete};
                
                \draw[arrow] (conftent.east) -- ($(conftent.east)+(1,0)$) node[midway, above] () {Yes} node[right] () {Mark Confirmed};
                
                \draw[arrow] (initnew.east) -- ($(initnew.east)+(1,0)$) node[right] () {Mark Tentative};
                
                \draw[arrow] ($(stage3.east)+(0, 0.2)$) -| (deltent) node[midway, near start, below]{Remaining Tracks};
                
                \draw[arrow] ($(stage3.east)+(0, -0.2)$) -- ($(stage3.east)+(0.2, -0.2)$) |- (initnew.west) node[midway, near end, below]{Remaining Detections};
                
                \draw[arrow] (stage3.south) |- (conftent.west) node[midway, near end, below](){Matched Tracks};
                
                \end{tikzpicture}
                }
                \caption{The association algorithm. At the start of the current time step, detections are obtained, and target positions are predicted with both the image and ground plane filters. Confirmed tracks are associated with high-confidence detections in the first association stage, which uses the association score in (\ref{stage1_score}). The remaining tracks are associated with the low-confidence detections and the remaining detections in the second association stage, which uses the association score in (\ref{stage2_score}). Tracks remaining after the second association stage are deleted if they are older than the threshold $\Omega$; otherwise, their states are changed to \textit{coasted}. The remaining high-confidence detections are associated with tentative tracks in the third association stage. If a tentative track is successfully matched with detections for two consecutive time steps, its state is changed to \textit{confirmed}; otherwise, it is deleted. The remaining high-confidence detections are used to initialise new tracks.}
                \label{fig:alg_flow}
            \end{figure*}

    \section{Experiments}\label{sec:exps}
        \noindent This section describes the experimental procedure used to find optimal hyperparameters for IMM-JHSE and reports the results of experiments on validation data. The datasets considered during these experiments are the MOT17 \citep{mot17} and DanceTrack \citep{dance} datasets. Results on the test sets are reported in Section~\ref{sec:results} and include the MOT20 \citep{mot20} and KITTI \citep{kitti} datasets. All tests and experiments are performed on a single core with a Ryzen 5900X CPU. Inference time is not a major concern in this paper, with optimisation being deferred to future work. Nevertheless, the current state achieves an average of 162.5 frames per second on the MOT17 dataset. Note that since publicly available detections are reused, this is not indicative of the expected inference time in a practical scenario.

        For evaluation, the TrackEval library is used \citep{trackeval} to obtain the higher order tracking accuracy (HOTA), association accuracy (AssA), IDF1 and multi-object tracking accuracy (MOTA). It has become standard to report these metrics in the most closely related recent work \citep{ucmc, Yang2022HardSpace}, and we refer the reader to the paper which introduces HOTA \citep{hota} for detailed discussions of these metrics, including the use of the Hungarian algorithm. Importantly, \citep{hota} clearly explains the numerous deficiencies of metrics like MOTA and IDF1 that are addressed by HOTA. As such, we consider HOTA the primary evaluation metric in this study. However, since HOTA can be decomposed into separate detection accuracy and association accuracy scores \citep{hota}, and publicly available detections are used presently, it is especially appropriate to compare AssA between methods that use the same detections. Since this work utilises publicly available detections, the detection accuracy (DetA) is not reported to avoid distracting from association performance.

    \subsection{Evaluation metrics}
        The MOT metrics used in this work are briefly introduced in this subsection.

        \subsubsection{Multiple object tracking accuracy}
            MOTA evaluates association accuracy by counting identity switches ($\mathrm{IDSW}$). However, it also takes into account detection false positives (FPs) and false negatives (FNs):
                \begin{equation*}
                    \text { MOTA }=1-\frac{|\mathrm{FN}|+|\mathrm{FP}|+|\mathrm{IDSW}|}{|\mathrm{gtDet}|},
                \end{equation*}
            where $\mathrm{gtDet}$ refers to ground truth detections. Here, $|\cdot|$ denotes the total number of the quantity concerned. In simple terms, MOTA adds the number of false positive and false negative detections to the number of identity switches, normalises the result by the number of ground truth detections and subtracts this from one \citep{hota}. 

        \subsubsection{IDF1}
            IDF1, or identity (ID) F1 score, considers entire track trajectories. Whereas MOTA counts identity switches once each time a track is assigned one ID in one frame and another in the next, IDF1 considers the proportion of overall time that a track is assigned the correct ID over its lifetime. IDF1 is calculated from ID true positives (IDTPs), ID false negatives (IDFNs) and ID false positives (IDFPs) \citep{hota}:
            \begin{equation*}
                \mathrm{IDF} 1=\frac{|\mathrm{IDTP}|}{|\mathrm{IDTP}|+0.5|\mathrm{IDFN}|+0.5|\mathrm{IDFP}|}.
            \end{equation*}

        \subsubsection{Higher-order tracking accuracy}
            HOTA \citep{hota} introduces the concept of true positive associations (TPAs), false negative associations (FNAs) and false positive associations (FPAs), in addition to detection-level true positives (TPs), FPs and FNs. At a localisation threshold $\alpha$, it is calculated as follows:
            \begin{equation*}
                \begin{aligned}
                & \operatorname{HOTA}_\alpha=\sqrt{\frac{\sum_{c \in\{\mathrm{TP}\}} \mathcal{A}(c)}{|\mathrm{TP}|+|\mathrm{FN}|+|\mathrm{FP}|}},\\
                & \mathcal{A}(c)=\frac{|\mathrm{TPA}(c)|}{|\operatorname{TPA}(c)|+|\operatorname{FNA}(c)|+|\mathrm{FPA}(c)|},
                \end{aligned}
            \end{equation*}
            where $\sum_{c \in\{\mathrm{TP}\}}$ indicates the sum over every true positive detection $c$ \citep{hota}. The final HOTA score is obtained by averaging $\operatorname{HOTA}_\alpha$ over various values of $\alpha$ \citep{hota}.
    
            HOTA can also be decomposed into different error types, such as localisation error, detection accuracy, detection precision, detection recall, association accuracy, association precision, and association recall \citep{hota}.

        \subsubsection{Association Accuracy}
            Association accuracy is calculated as follows:
            \begin{equation*}
                \begin{aligned}
                &\operatorname{AssRe}_\alpha=\frac{1}{|\mathrm{TP}|} \sum_{c \in\{\mathrm{TP}\}} \frac{|\mathrm{TPA}(c)|}{|\mathrm{TPA}(c)|+|\operatorname{FNA}(c)|}\\
                &\operatorname{AssPr}_\alpha=\frac{1}{|\mathrm{TP}|} \sum_{c \in\{\mathrm{TP}\}} \frac{|\operatorname{TPA}(c)|}{|\operatorname{TPA}(c)|+|\operatorname{FPA}(c)|}\\
                &\operatorname{AssA}_\alpha=\frac{\operatorname{AssRe}_\alpha \cdot \operatorname{AssPr}_\alpha}{\operatorname{AssRe}_\alpha+\operatorname{AssPr}_\alpha-\operatorname{AssRe}_\alpha \cdot \operatorname{AssPr}_\alpha}.
                \end{aligned}
            \end{equation*}
            Like HOTA, the final AssA score is obtained by averaging $\operatorname{AssA}_\alpha$ over various values of $\alpha$. Association precision and recall are also novel contributions of \citep{hota}: association recall will be low if a tracker splits a ground truth track into multiple tracks, and association precision will be low when a tracker predicts the same identity for multiple objects. Since IMM-JHSE uses publicly available detections, it is appropriate to consider association accuracy and IDF1 specifically in comparison with other methods that use the same detections.
        
    \subsection{Parameter selection}
        The largest assumption of IMM-JHSE is that an initial homography estimate is available. For all experiments and test set results, the camera calibration matrices provided in \citep{ucmc} are used -- except for the KITTI dataset, which provides its own. Furthermore, the estimates of $\mathbf{\hat{A}}_t$ (\ref{affine_mat}) provided by \citep{ucmc} are used for the MOT17 and MOT20 datasets, but for the DanceTrack and KITTI datasets $\mathbf{\hat{A}}_t$ is estimated as mentioned in Section~\ref{sec:back}.
    
        For IMM-JHSE, the following hyperparameters need to be specified:
        $\sigma_x, \sigma_y, \alpha_1, \alpha_2, \alpha_3, \Omega, b, d_{\text{high}}, d_{\text{low}}$, as well as the transition probabilities illustrated in Figure~\ref{fig:method_graphical}. For the transition probabilities, it is only necessary to specify the self-transition probabilities $p_{s,s}, p_{d,d}, p_{W,W}$ and $p_{I,I}$, since the other transition probabilities can be derived from them. 

        \subsubsection{Filter initialisation}
        In addition to the parameters above, the covariance matrices with which each filter is initialised must be specified. The initial covariance matrix associated with the ground plane coordinates $\mathbf{x}^W_t$ is obtained similar to \citep{ucmc}:
            \begin{equation}
                \mathbf{J^{IW}_t}\mathbf{R}_0\left(\mathbf{J^{IW}_t}\right)^T,
            \end{equation}
        where $\mathbf{J^{IW}_t}$ is obtained from (\ref{jac_g}) and $\mathbf{R}_0$ from (\ref{initial_r0}). The variance in each velocity component is set to $v$, which is determined with the pattern search algorithm as described subsequently. Finally, the initial covariance associated with the homography elements is set to $\mathbf{0}$. Thus, a perfect homography estimate is assumed initially, but this covariance is dynamically adjusted by (\ref{homog_dyn_noise}). 
        
        This paper uses the pattern search algorithm \citep{pattern_search} to find optimum values for the hyperparameters. For each experiment described subsequently, the initial values are set as follows: $\sigma_x=5, \sigma_y=5, \alpha_1=\alpha_2=\alpha_3=0.5, \Omega=30, b=0, d_{\text{high}}=0.6, d_{\text{low}}=0.5, p_{s,s}=p_{d,d}=p_{W,W}=p_{I,I}=0.9,v=0.5$, and the number of pattern search iterations is restricted to 200. The objective function selected for optimisation is HOTA for the MOT17, MOT20 and KITTI datasets and association accuracy (AssA) for the DanceTrack dataset. Other evaluation metrics include IDF1 and multiple object tracking accuracy (MOTA). The reader is referred to \citep{hota} for details regarding these metrics. In all cases, evaluation is performed with the TrackEval library \citep{trackeval}. For the DanceTrack, KITTI and MOT20 datasets, the average of the objective functions of each video in the entire dataset is optimised, while for the MOT17 dataset, the objective function is optimised per video.

        The parameters obtained for IMM-JHSE with the pattern search algorithm are given in Table~\ref{params}. On average, the pattern search algorithm takes 1.2 hours to find the best parameters on each dataset.

\subsection{Ablation study on the DanceTrack dataset}
        Table~\ref{tab:dance_ablation} reports the results of an ablation study on the DanceTrack validation dataset. The proposed method is focused on handling the situation where targets regularly leave the ground plane. As such, we do not perform ablation studies on other datasets. In light of this, a deficiency of our proposed method is that it is designed for the DanceTrack dataset in particular. 
        
        The second column in Table~\ref{tab:dance_ablation} refers to using the dynamic measurement noise estimation method described in Section~\ref{noise_est}. When dynamic measurement noise is not used (as indicated with \text{\sffamily X}), (\ref{initial_r0}) is used to provide an estimate of the measurement noise at each time step. When the dynamic estimation method is not used, HOTA decreases from $63.81$ to $60.80$. Thus, the dynamic noise estimation method contributes significantly to the success of IMM-JHSE. 
        
        The third column refers to the static camera motion model (\ref{homog_static}). IMM-JHSE is restricted to the dynamic camera motion model of (\ref{homog_dyn}) without the static camera motion model. In this case, HOTA decreases to $62.04$ -- showing that the static camera motion model (\ref{homog_static}) is beneficial. 
        
        The fourth column refers to using the dynamic camera motion model (\ref{homog_dyn}). Without this model, HOTA decreases to $61.19$. Therefore, the dynamic camera motion model plays a more significant role than the static camera motion model. 
        
        The fifth column refers to using the image plane filter and the BIoU. Without these, all association scores are calculated by $P(D)d_{\text{conf}}$, and HOTA decreases to $63.03$. Thus, using bounding box information with the ground plane state vector proves beneficial.  

        The best results are obtained in the first row when all of the components of IMM-JHSE are in use.
        \begin{table*}[h]
            \centering
            \newcolumntype{C}{>{\centering\arraybackslash} p{2cm} }
            \begin{tabular}{CCCCCCC}
            \hline
                Method & Adaptive Measurement Noise & Static Camera Motion Model &  Dynamic Camera Motion Model & Bounding Box Filter & HOTA$\uparrow$ & IDF1$\uparrow$ \\\hline
                    IMM-JHSE (Ours) & \checkmark & \checkmark & \checkmark & \checkmark & \textbf{63.81} & \textbf{69.43}\\\hline
                    IMM-JHSE (Ours) & \checkmark & \checkmark & \checkmark & \text{\sffamily X} & 63.03 & 67.14\\\hline
                    IMM-JHSE (Ours) & \checkmark & \checkmark & \text{\sffamily X} & \checkmark & 61.19 & 65.69\\\hline
                    IMM-JHSE (Ours) & \checkmark & \text{\sffamily X} & \checkmark & \checkmark & 62.04 & 67.59\\\hline
                    IMM-JHSE (Ours) & \text{\sffamily X} & \checkmark & \checkmark & \checkmark & 60.80 & 64.17\\\hline
                    UCMC \citep{ucmc} & \text{\sffamily X} & \text{\sffamily X} & \text{\sffamily X} & \text{\sffamily X} &  60.42 & 62.64\\\hline
                    UCMC+ \citep{ucmc} & \text{\sffamily X} & \text{\sffamily X} & \text{\sffamily X} & \text{\sffamily X} & 59.18 & 62.52\\\hline
            \end{tabular}
            \caption{Ablation study on the DanceTrack validation set, with detections obtained from \citep{Cao2022Observation-CentricTracking}. The columns denote various components of IMM-JHSE as described in Section~\ref{sec:exps}. UCMC+ refers to the method of \citep{ucmc} with camera motion compensation.}
            \label{tab:dance_ablation}
        \end{table*}

    \subsection{Investigating association score mixture methods}
\begin{figure*}[!t]
                    \centering
                    \subfloat[While the performers are on the ground, the ground plane and image plane filter probabilities are more or less equal.]{\includegraphics[width=0.45\textwidth]{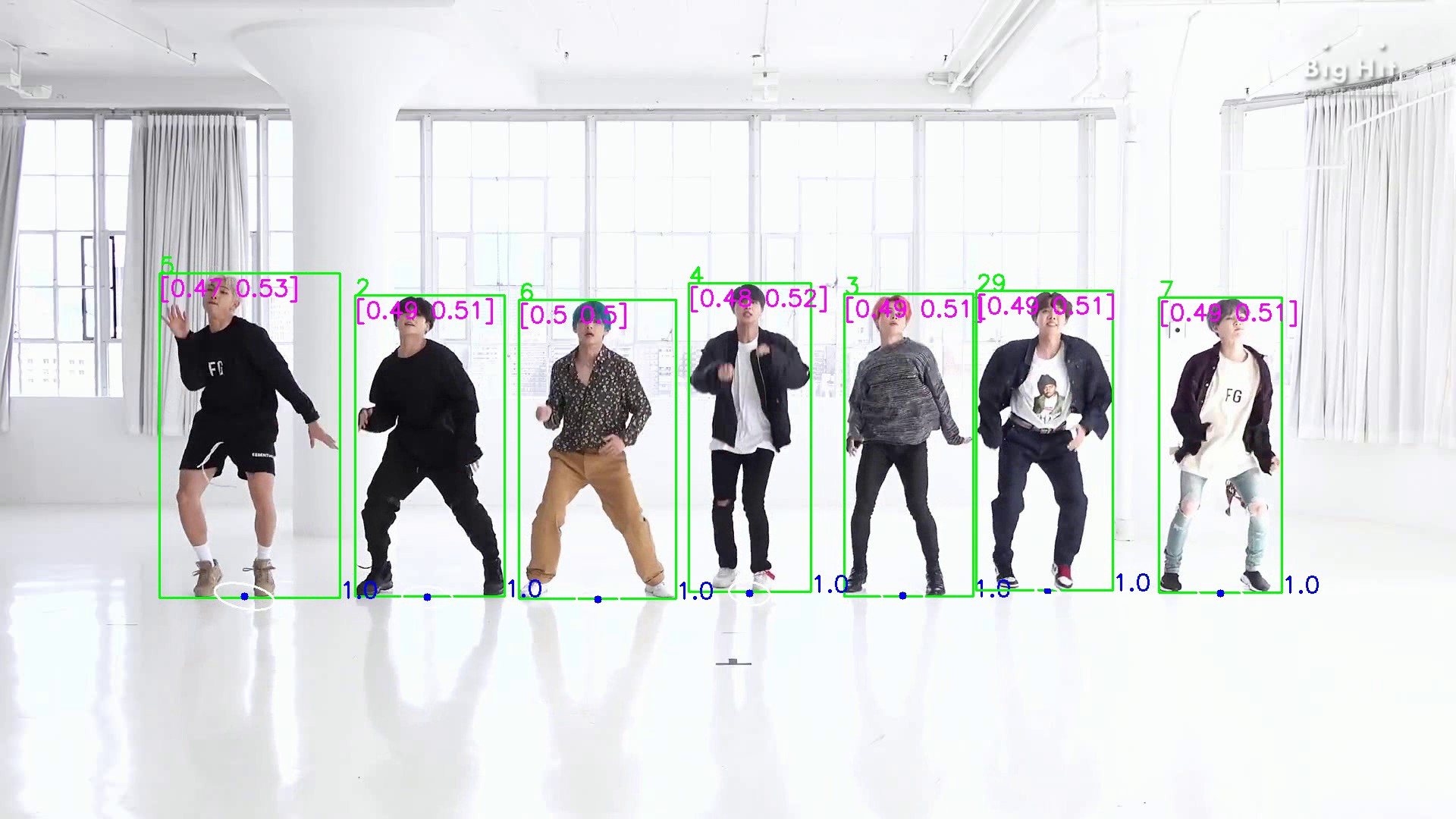}%
                    \label{fig:jump_a}}
                    \hfil
                    \subfloat[As the performers start to jump, the image plane filter probabilities become larger than the ground plane filter probabilities.]{\includegraphics[width=0.45\textwidth]{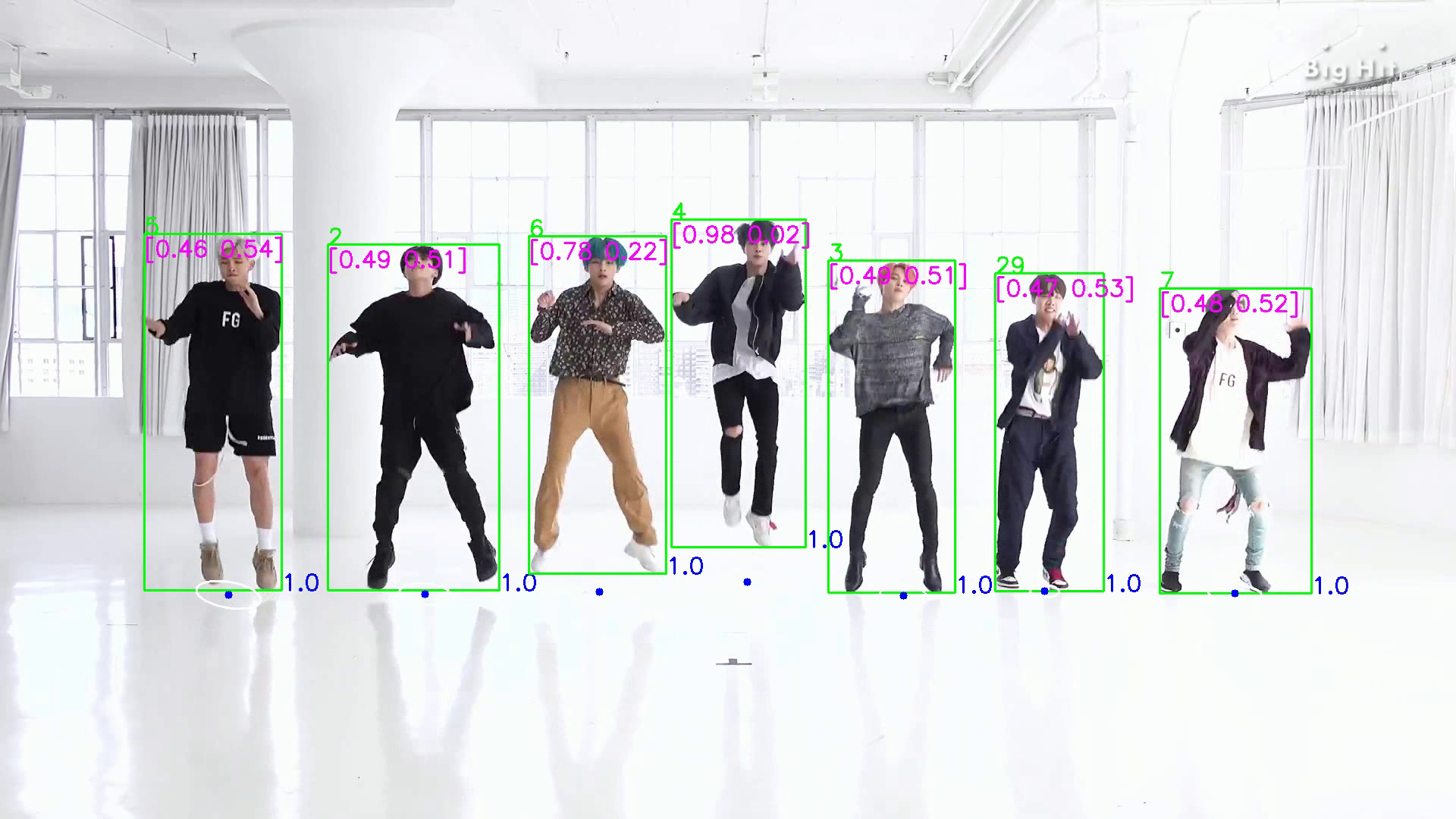}%
                    \label{fig:jump_b}}
                    \hfil
                    \subfloat[For targets that have left the ground plane, the image plane filter probabilities are much larger than those of the ground plane filter, and association is dominated by the BIoU (\ref{stage2_score}).]{\includegraphics[width=0.45\textwidth]{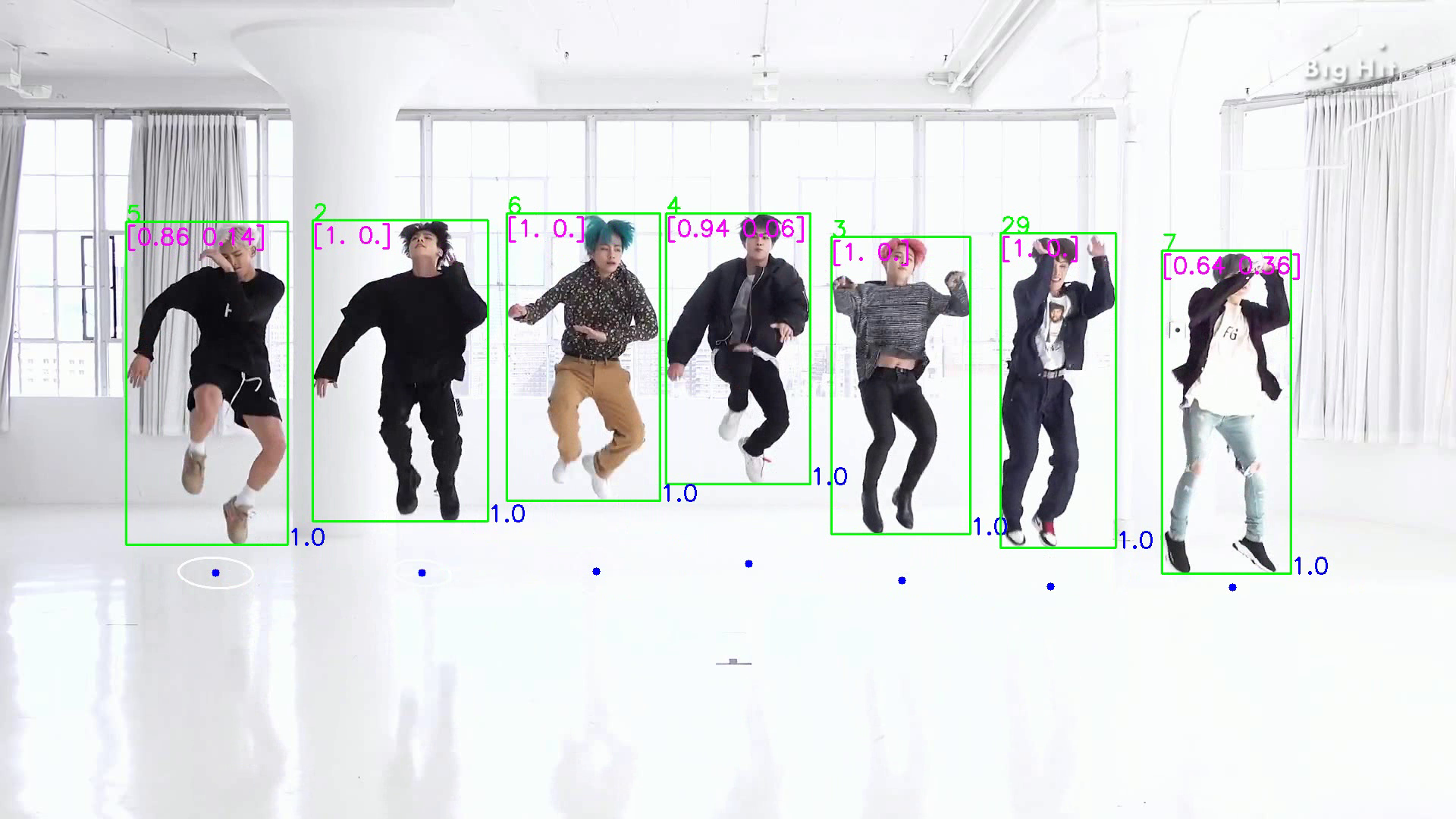}%
                    \label{fig:jump_c}}
                    \caption{Showing that the association score in (\ref{stage2_score}) allows track identities to be maintained while targets leave the ground plane. The list displayed in purple represents the model probabilities of the image plane and ground plane filters, respectively, for each track. The image plane filter dominates the total model likelihood as targets leave the ground plane.}
                    \label{fig:jump}
                \end{figure*}

        In Table~\ref{tab:bbox_ablation}, the way bounding box information is considered is varied. For ``IMM-like'', the association score for stages 2 and 3 remains as in (\ref{stage2_score}). For ``Multiplicative'', all stages use the association score in (\ref{stage1_score}). The ``IMM-like'' association scores achieve the best performance across all metrics for MOT17 and DanceTrack. Furthermore, the benefit of using the ``IMM-like'' association score is illustrated in Figure~\ref{fig:jump}, where track identities are maintained while the performers jump, and the ground plane position estimates cannot be relied upon. Thus, the proposed association score in (\ref{stage2_score}) is justified.
        \begin{table}[h]
            \centering
            \small
            \newcolumntype{C}{>{\centering\arraybackslash} p{0.125\columnwidth} }
            \begin{tabular}{CCCCCC}
            \hline
                Dataset & Bounding Box Score & HOTA $\uparrow$ & IDF1 $\uparrow$ & MOTA $\uparrow$ & AssA $\uparrow$\\\hline

                \multirow{2}{*}{MOT17} & IMM-like & \textbf{75.96} & \textbf{88.55} & \textbf{86.97} & \textbf{77.60}\\
                & Multi-plicative & 75.62 & 87.71 & 86.43 & 77.25\\\hline

                \multirow{2}{*}{DanceTrack} & IMM-like & \textbf{63.81} & \textbf{69.43} & \textbf{85.87} & \textbf{53.65}\\
                & Multi-plicative & 61.71 & 66.27 & 84.59 & 50.72\\\hline
                
            \end{tabular}
            \caption{Validation results on the MOT17 and DanceTrack training sets with detections obtained from \citep{Zhang2022ByteTrack:Box}, where the way in which bounding box information is used is varied.}
            \label{tab:bbox_ablation}
        \end{table}

    \subsection{Investigation of the chosen objective function}
        Table~\ref{tab:optim_metric} shows that for some datasets, such as the DanceTrack dataset, it may be beneficial to optimise for AssA instead of HOTA. Conversely, optimising for AssA for the MOT17 dataset drastically reduces HOTA, IDF1 and MOTA. A possible explanation is that partial detections (e.g. slightly occluded tracks) are correctly associated. As a result, AssA increases, but localisation accuracy may decrease since the associated detection is not well aligned with the ground truth. This hints that specific datasets (such as MOT17) could benefit significantly from increased detector localisation accuracy.
        Furthermore, note that all metrics significantly improve for the DanceTrack dataset when the objective function is optimised per video instead of optimising the combined objective function over the entire dataset. Nevertheless, for the DanceTrack, MOT20 and KITTI datasets, the parameters optimised on the combined objective function are used for testing in the subsequent section. This is to avoid estimating an association between validation and testing videos. However, for the MOT17 dataset, this association is clear and manageable since it consists of fewer videos. Thus, for MOT17, the parameters optimised per video are used.
        \begin{table}[h]
            \centering
            \small
            \newcolumntype{C}{>{\centering\arraybackslash} m{0.125\columnwidth} }
            \begin{tabular}{CCCCCC}
            \hline
                Dataset & Optimised Metric & HOTA $\uparrow$ & IDF1 $\uparrow$ & MOTA $\uparrow$ & AssA $\uparrow$\\\hline

                \multirow{2}{*}{MOT17} & HOTA & \textbf{75.96} & \textbf{88.55} & \textbf{86.97} & 77.6\\
                & AssA & 70.69 & 80.07 & 69.9 & \textbf{81.21}\\\hline

                \multirow{3}{*}{DanceTrack} & HOTA & 63.14 & 68.19 & 84.67 & 53.21\\
                & AssA & 63.81 & 69.43 & 85.87 & 53.65\\
                 & AssA$^*$ & \textbf{70.14} & \textbf{76.23} & \textbf{86.79} & \textbf{63.88}\\\hline
                 \multicolumn{6}{l}{\footnotesize $^*$ Optimised per video.}
                \\
                
            \end{tabular}
            \caption{Validation results on the MOT17 and DanceTrack training sets with detections obtained from \citep{Zhang2022ByteTrack:Box}, where the optimised metric is varied. Note that, for some datasets, the results depend greatly on the metric being optimised.}
            \label{tab:optim_metric}
        \end{table}
        
    \section{Results}\label{sec:results}    
        \noindent This section presents the results of IMM-JHSE evaluated on the DanceTrack, MOT17, MOT20 and KITTI test datasets. Note that the source of detections is indicated in the table captions. When comparing to tracking-by-detection methods, the same publicly available detections are used. Thus, enabling a fair comparison between the association methods only. For attention and 3D MOT methods, we refer the reader to the relevant citations for information about the vast variety of detectors and training strategies.
        
        \subsection{Results on DanceTrack}
        The results for the DanceTrack test set are shown in Table~\ref{tab:dance_test}, where IMM-JHSE significantly outperforms related methods (ByteTrack, C-BIoU, OC-SORT and UCMCTrack) in terms of HOTA, IDF1 and AssA, using the same detections. Specifically, compared to the runner-up (UCMCTrack+), HOTA increases by $2.64$, IDF1 by $6.72$ and AssA by $4.11$. Following the ablation study presented in Table~\ref{tab:dance_ablation}, this is attributed to IMM-JHSE's dynamic noise estimation, ground plane IMM filter, which accounts for two camera motion models, and its specific use of BIoU during association. Note that MOTA is the only metric that has not improved over previous methods. This may be because tentative tracks are removed from the result file if they are never confirmed. Thus increasing the number of false negative detections and reducing the MOTA. It is also possible that partial detections are associated with IMM-JHSE, which do not have corresponding ground truth detections because the object is, for example, mostly occluded. This would increase the number of false positive detections, which could also cause a decrease in MOTA. 
        
        IMM-JHSE performs remarkably well compared to attention-based methods. Despite FusionTrack's ability to perform joint detection, tracking, and exploit appearance features, IMM-JHSE outperforms it in terms of HOTA. IMM-JHSE performs similarly to the computationally expensive MeMOTR with Deformable DETR, even outperforming it with IDF1 and association accuracy. IMM-JHSE also outperforms MeMOTR using DAB-Deformable DETR in IDF1. Considering the vast difference between the methods of IMM-JHSE and MOTIP with Deformable DETR, it is remarkable that IMM-JHSE obtains similar performance in most metrics. However, the advantage of the end-to-end approach proves to be great with the combination of MOTIP and DAB-Deformable DETR, which significantly outperforms IMM-JHSE. 
        
        IMM-JHSE uses a single initial homography estimate and maintains the state of the homography projection matrix separately for each track. This strategy may aid association by allowing individual track homography estimates to diverge from the unobserved universal transformation, thus serving as an additional source of disambiguation between tracks. In contrast, attention-based methods are at a disadvantage in this regard since they do not incorporate ground-plane information.
        \begin{table}[htb!]
            \centering
            \newcolumntype{C}{>{\centering\arraybackslash} m{0.15\columnwidth} }
            \begin{tabular}{CCCCC}
            \hline
                Method & HOTA $\uparrow$ & IDF1 $\uparrow$ & MOTA $\uparrow$ & AssA $\uparrow$\\\hline
                \multicolumn{5}{p{0.75\columnwidth}}{Tracking-by-detection} \\\hline
                ByteTrack \citep{Zhang2022ByteTrack:Box} & 47.3 & 52.5 & 89.5 & 31.4\\\hline
                C-BIoU \citep{Yang2022HardSpace} & 60.6 & 61.6 & 91.6 & 45.4\\\hline
                OC-SORT \citep{Cao2022Observation-CentricTracking} & 55.1 & 54.9 & \textbf{92.2} & 40.4\\\hline
                UCMCTrack \citep{ucmc} & 63.4 & 65.0 & 88.8 & 51.1\\\hline
                UCMCTrack+ \citep{ucmc}& 63.6 & 65.0 & 88.9 & 51.3\\\hline
                \textbf{IMM-JHSE} (Ours) & \textbf{66.24} & \textbf{71.72} & 89.95 & \textbf{55.41}\\\hline
                \multicolumn{5}{p{0.75\columnwidth}}{Tracking-by-attention} \\\hline
                FusionTrack \citep{fusiontrack} & 65.3 & 73.3 & 90.1 & 57.5\\\hline
                MeMOTR$^*$ \citep{Gao2024MeMOTR:Tracking} & 66.7 & 70.6 & \textbf{94} & 53\\\hline
                MeMOTR$^\dagger$ \citep{Gao2024MeMOTR:Tracking} & 68.5 & 71.2 & 89.9 & 58.4\\\hline
                MOTIP$^*$ \citep{Gao2024MultiplePrediction} & 67.5 & 72.2 & 90.3 & 57.6\\\hline
                MOTIP$^\dagger$ \citep{Gao2024MultiplePrediction} & \textbf{70} & \textbf{75.1} & 91 & \textbf{60.8}\\\hline
                \multicolumn{5}{p{0.75\columnwidth}}{$^*$ Deformable DETR \citep{Zhu2021DeformableDetection}}\\
                \multicolumn{5}{p{0.75\columnwidth}}{$^\dagger$ DAB-Deformable DETR \citep{Liu2022DAB-DETR:DETR}}\\
                
            \end{tabular}
            \caption{Results on the DanceTrack test set. For the tracking-by-detection methods, the detections are obtained from \citep{Zhang2022ByteTrack:Box}.}
            \label{tab:dance_test}
        \end{table}
    \subsection{Results on the KITTI dataset}\label{kittiresults}
        For the KITTI test dataset, IMM-JHSE significantly outperforms the HOTA and AssA scores of other methods (UCMCTrack and OC-SORT) in the car class -- with HOTA increasing by $2.11$ over UCMCTrack and AssA by $5.09$ in Table~\ref{tab:kitti_test}. 
        
        At the time of writing, IMM-JHSE is placed twenty-fourth on the KITTI-car leaderboard\footnote{\url{https://www.cvlibs.net/datasets/kitti/eval_tracking_detail.php?result=c3e8939f18cba7b23a2e05a92d4df9e994eaaf91}}. In addition, all methods that are placed above it are 3D MOT methods that make use of regular 3D measurements, except for RAM \citep{Tokmakov2022ObjectMemory}, which uses a self-supervised encoder that optimises a Markov walk along a space-time graph. Table~\ref{tab:kitti_test_3d} compares IMM-JHSE and some of these methods. Interestingly, while RAM outperforms IMM-JHSE in the car class, IMM-JHSE outperforms RAM in the pedestrian class. While there is a large difference (3.49 HOTA) between IMM-JHSE and the top-ranked method (BiTrack \citep{Huang2024BiTrack:Data}), it should also be noted that IMM-JHSE uses publicly available detections, and the corresponding difference in detection accuracy is 3.64\%. Furthermore, BiTrack is an offline method. Considering that IMM-JHSE uses only an initial homography estimate and further relies only on 2D detections, it is remarkable that it compares so well to 3D MOT methods (which use regular 3D measurements). This implies that the additional disambiguating information afforded by 3D methods is closely matched by the expanded track state maintained in IMM-JHSE and its novel association method. Practically, the use of IMM-JHSE foregoes processing 3D measurements in every frame, although this comes at the cost of a more complex state and association model.
        
        In Table~\ref{tab:kitti_test}, which consists of the top three KITTI-pedestrian methods, IMM-JHSE shows the poorest results for the pedestrian class in all metrics except AssA (though it still performs well; it is ranked second on the online KITTI-pedestrian leaderboard --- UCMCTrack's results do not appear there). As will be seen in Table~\ref{tab:mot17_test} and Table~\ref{tab:mot20_test} for the MOT17 and MOT20 datasets, IMM-JHSE performs worse with tracks that take up relatively little of the image plane area, particularly if a multitude of such tracks are close to one another. In other words, the merits of IMM-JHSE are more evident when tracking relatively large (in terms of bounding box area) objects that move irregularly or with high acceleration or velocity in the image plane rather than in highly cluttered situations. This may be due to the second association stage, which is designed to account for ambiguous detections that may diverge from that expected by the predicted ground plane state. Yet, suppose multiple tracks are close to one another, and all have ambiguous detections. In that case, it is conceivable that they may be confused with one another since they are likely to have high BIoU scores with detections that belong to other tracks.

\begin{figure}[!htbp]
                \centering
                      \begin{minipage}[b]{\linewidth}
                      \centering\includegraphics[width=\textwidth]{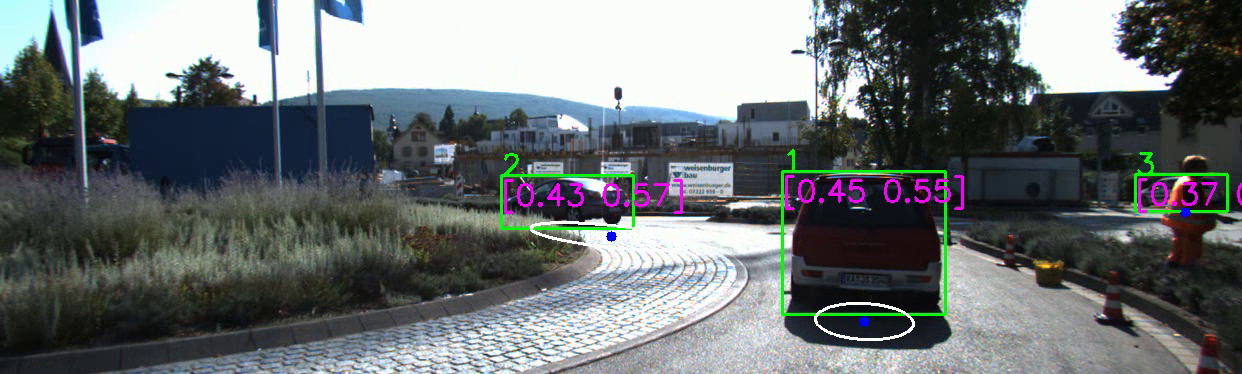}
                      \subcaption{}\label{fig:noped1}
                    \end{minipage}%
                    \hfill
                    \begin{minipage}[b]{\linewidth}
                      \centering\includegraphics[width=\textwidth]{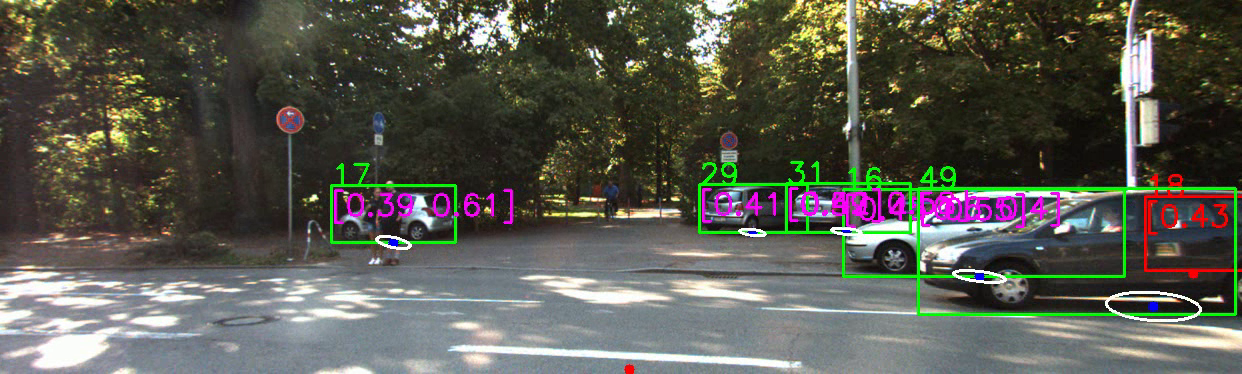}
                      \subcaption{}\label{fig:noped2}
                    \end{minipage}%
                \caption{This figure shows two examples where pedestrians are not detected by the PermaTrack detector.\\\subref{fig:noped1} The official on the right-hand side is undetected. \subref{fig:noped2} The couple in front of the car with ID 17 is undetected.}
                \label{fig:nopedestrian}
            \end{figure}

        It is also possible that pedestrians are not reliably detected by the PermaTrack detector. Figure~\ref{fig:nopedestrian} shows two such examples. Figure~\ref{fig:nopedestrian} also shows a side-effect of the dynamic noise estimation if the initial position or velocity variance is not set high enough: in both subfigures, there are tracks with relatively large uncertainty ellipses. This is because, for the KITTI dataset, the initial velocity variance is very small (see Table~\ref{params}). Therefore, when a fast-moving track is initialised, IMM-JHSE incorporates the prediction error into the process noise. In these situations, where tracks do not leave the ground plane, explicitly mixing ground plane and bounding box state estimates may be better.

\begin{figure}[!htbp]
        \centering
        \includegraphics[width=\linewidth]{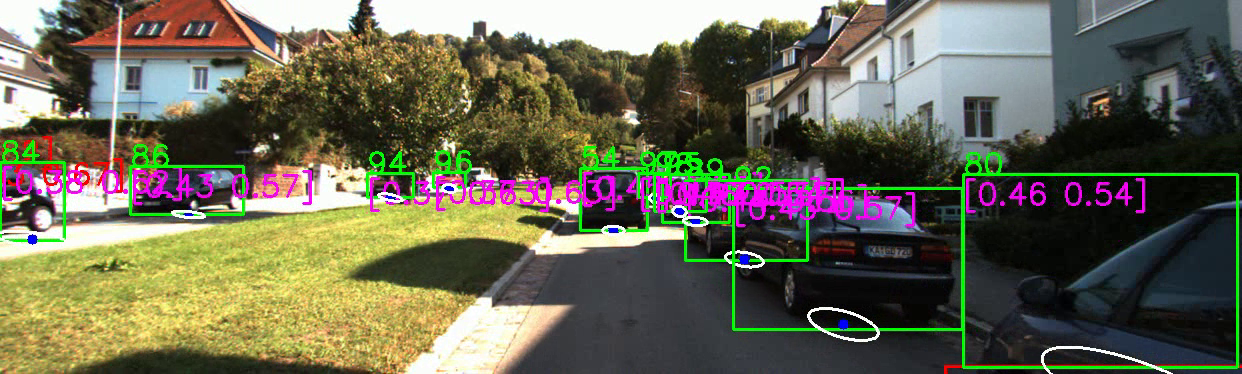}
        \caption{An example of the large number of car labels relative to pedestrian labels in the KITTI dataset.}
        \label{fig:kitticardom}
    \end{figure}

        Another possibility is that the optimised parameters overfit the validation set on the car class, reducing performance on the pedestrian test set. Indeed, the pattern search algorithm seems prone to getting stuck in local minima based on close observation during optimisation iterations. This seems especially likely since the KITTI dataset consists primarily of car labels. Figure~\ref{fig:kitticardom} shows an example of a sequence with few pedestrians.

        Despite its relatively poor performance with the pedestrian class, IMM-JHSE is placed third on the KITTI-pedestrian leaderboard at the time of writing (including UCMCTrack, which does not appear on the online leaderboard), where OC-SORT is the only published method that outperforms it.
        
        \begin{table*}[ht!]
            \centering
            \newcolumntype{C}{>{\centering\arraybackslash} m{0.25\columnwidth} }
            \begin{tabular}{C|CCC|CCC}
            \hline
                Method & \multicolumn{3}{c|}{Car} & \multicolumn{3}{c}{Pedestrian}\\
                & HOTA $\uparrow$ & MOTA $\uparrow$ & AssA $\uparrow$& HOTA $\uparrow$ & MOTA $\uparrow$ & AssA $\uparrow$\\\hline

                OC-SORT \citep{Cao2022Observation-CentricTracking} & 76.5 & 90.3 & 76.4 & 54.7 & 65.1 & \textbf{59.1}\\\hline
                UCMCTrack \citep{ucmc} & 77.1 & \textbf{90.4} & 77.2 & \textbf{55.2} & \textbf{67.4} & 58.0\\\hline
                UCMCTrack+ \citep{ucmc}& 74.2 & 90.2 & 71.7 & 54.3 & 67.2 & 56.3\\\hline
                \textbf{IMM-JHSE} (Ours) & \textbf{79.21} & 89.8 & \textbf{82.29} & 54.07 & 64.95 & 56.89\\\hline
                
            \end{tabular}
            \caption{Results on the KITTI test set with detections obtained from PermaTrack \citep{permatrack}.}
            \label{tab:kitti_test}
        \end{table*}
        \begin{table*}[ht!]
            \centering
            \newcolumntype{C}{>{\centering\arraybackslash} m{0.25\columnwidth} }
            \begin{tabular}{C|CCC|CCC}
            \hline
                Method & \multicolumn{3}{c|}{Car} & \multicolumn{3}{c}{Pedestrian}\\
                & HOTA $\uparrow$ & MOTA $\uparrow$ & AssA $\uparrow$& HOTA $\uparrow$ & MOTA $\uparrow$ & AssA $\uparrow$\\\hline

                AB3DMOT\\+PointRCNN \citep{Weng20203DMetrics} & 69.99 & 83.61 & 69.33 & 37.81 & 38.13 & 44.33\\\hline

                Mono 3D KF \citep{Reich2021MonocularTracks} & 75.47 & 88.48 & 77.63 & 42.87 & 45.44 & 46.31\\\hline
                
                OC-SORT \citep{Cao2022Observation-CentricTracking} & 76.54 & 90.28 & 76.39 & \textbf{54.69} & 65.14 & \textbf{59.08}\\\hline

                PermaTrack \citep{permatrack} & 78.03 & 91.33 & 78.41 & 48.63 & 65.98 & 45.61\\\hline
                
                FusionTrack\\+pointgnn \citep{fusiontrack} & 78.50 & 89.57 & 81.89 & -- & -- & --\\\hline

                \textbf{IMM-JHSE} (Ours) & 79.21 & 89.8 & 82.29 & 54.07 & 64.95 & 56.89\\\hline
                
                RAM \citep{Tokmakov2022ObjectMemory} & 79.53 & 91.61 & 80.94 & 52.71 & \textbf{68.4} & 52.19\\\hline
                
                MCTrack (online) \citep{Wang2024MCTrack:Driving}& 80.78 & 89.82 & 84.30 & -- & -- & --\\\hline

                CasTrack \citep{Wu2022CasA:Clouds}& 81.00 & \textbf{91.91} & 84.22 & -- & -- & --\\\hline
                
                MCTrack (offline) \citep{Wang2024MCTrack:Driving}& 82.56 & 91.62 & \textbf{86.64} & -- & -- & --\\\hline
                
                BiTrack (offline) \citep{Huang2024BiTrack:Data}& \textbf{82.7} & 91.55 & 86.17 & -- & -- & --\\\hline
                
            \end{tabular}
            \caption{A comparison of IMM-JHSE to other methods on the KITTI leaderboard.}
            \label{tab:kitti_test_3d}
        \end{table*}

    \subsection{Results on the MOT datasets}
        IMM-JHSE achieves the results reported in Table~\ref{tab:mot17_test} on the MOT17 test dataset\footnote{\url{https://motchallenge.net/method/MOT=8468&chl=10}}. While IMM-JHSE does not achieve the best result for any metric, it places itself firmly in between UCMCTrack and UCMCTrack+ for all metrics---outperforming all other methods besides UCMCTrack+---including ByteTrack, C-BIoU and OC-SORT---in HOTA, IDF1 and AssA. 

\begin{figure}[!htbp]
            \centering
            \includegraphics[width=\linewidth]{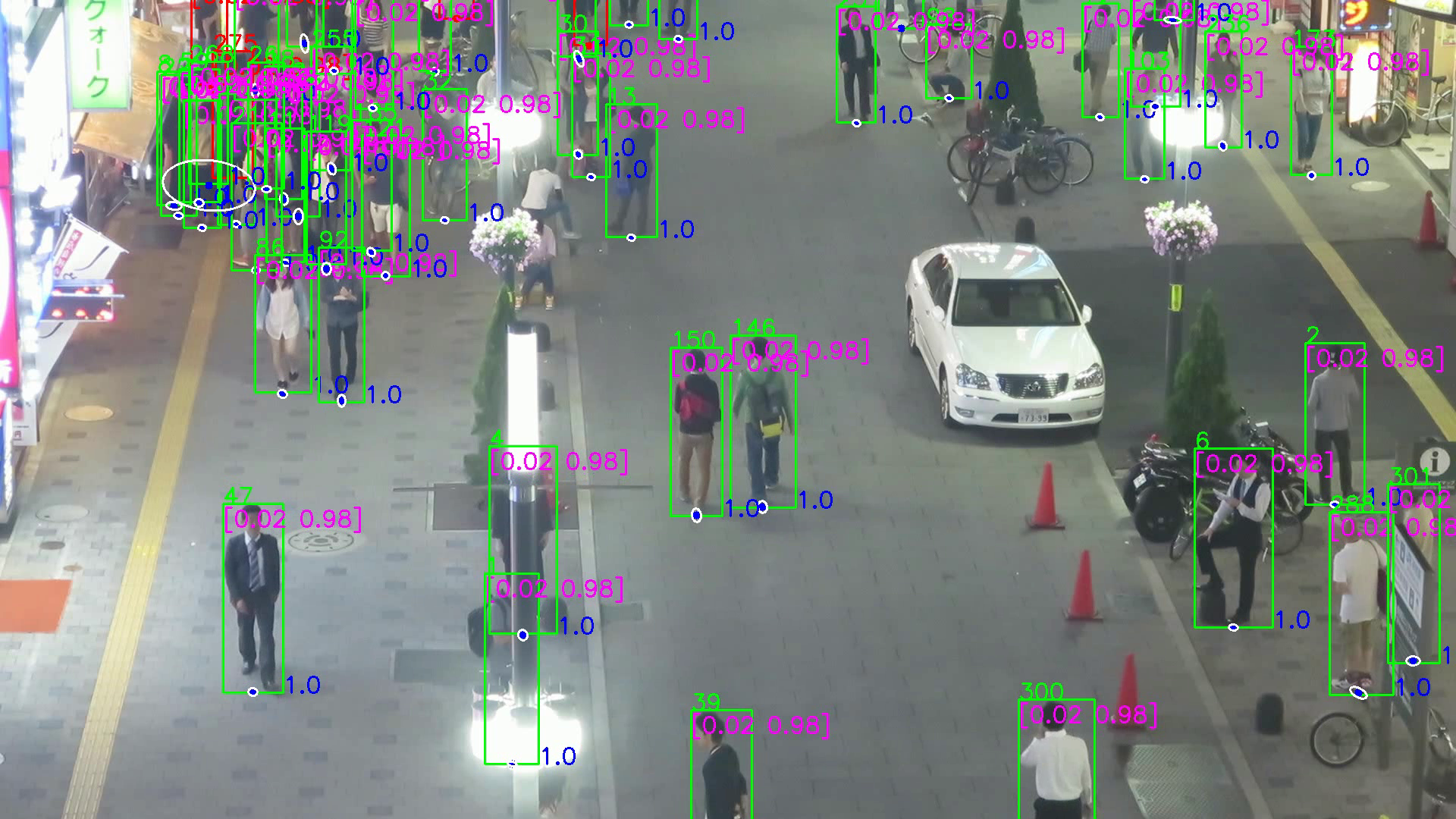}
            \caption{An example of a high track-density frame from the MOT17 dataset.}
            \label{fig:mot17crowd}
        \end{figure}

        IMM-JHSE also significantly outperforms tracking-by-attention methods on the MOT17 dataset. It is clear that the high track densities in the MOT datasets also present a challenge to tracking-by-attention methods. Figure~\ref{fig:mot17crowd} shows an example of a relatively high track-density frame from the MOT17 dataset. Notice that the ground plane position uncertainty is very high for a particular track in the track cluster on the left. This follows the discussion in Section~\ref{kittiresults} that partial detections can cause ambiguity in IMM-JHSE.
        \begin{table}[h]
            \centering
            \newcolumntype{C}{>{\centering\arraybackslash} m{0.155\columnwidth} }
            \begin{tabular}{CCCCC}
            \hline
                Method & HOTA $\uparrow$ & IDF1 $\uparrow$ & MOTA $\uparrow$ & AssA $\uparrow$\\\hline
                \multicolumn{5}{p{0.75\columnwidth}}{Tracking-by-detection} \\\hline
                ByteTrack \citep{Zhang2022ByteTrack:Box} & 63.1 & 77.3 & 80.3 & 62.0\\\hline
                C-BIoU \citep{Yang2022HardSpace} & 64.1 & 79.7 & \textbf{81.1} & 63.7\\\hline
                OC-SORT \citep{Cao2022Observation-CentricTracking} & 63.2 & 77.5 & 78.0 & 63.4\\\hline
                UCMCTrack \citep{ucmc} & 64.3 & 79.0 & 79.0 & 64.6\\\hline
                UCMCTrack+ \citep{ucmc}& \textbf{65.8} & \textbf{81.1} & 80.5 & \textbf{66.6}\\\hline
                \textbf{IMM-JHSE} (Ours) & 64.90 & 80.11 & 79.54 & 65.65\\\hline
                \multicolumn{5}{p{0.75\columnwidth}}{Tracking-by-attention} \\\hline
                MeMOTR$^*$ \citep{Gao2024MeMOTR:Tracking} & 58.8 & 71.5 & 72.8 & 58.4\\\hline
                MOTIP$^*$ \citep{Gao2024MultiplePrediction} & 59.2 & 71.2 & 75.5 & 56.9\\\hline
                \multicolumn{5}{p{0.75\columnwidth}}{$^*$ Deformable DETR \citep{Zhu2021DeformableDetection}}\\
            \end{tabular}
            \caption{Results on the MOT17 test set. For the tracking-by-detection methods, the detections are obtained from \citep{Zhang2022ByteTrack:Box}.}
            \label{tab:mot17_test}
        \end{table}
        
\begin{figure}[!htbp]
        \centering
        \includegraphics[width=\linewidth]{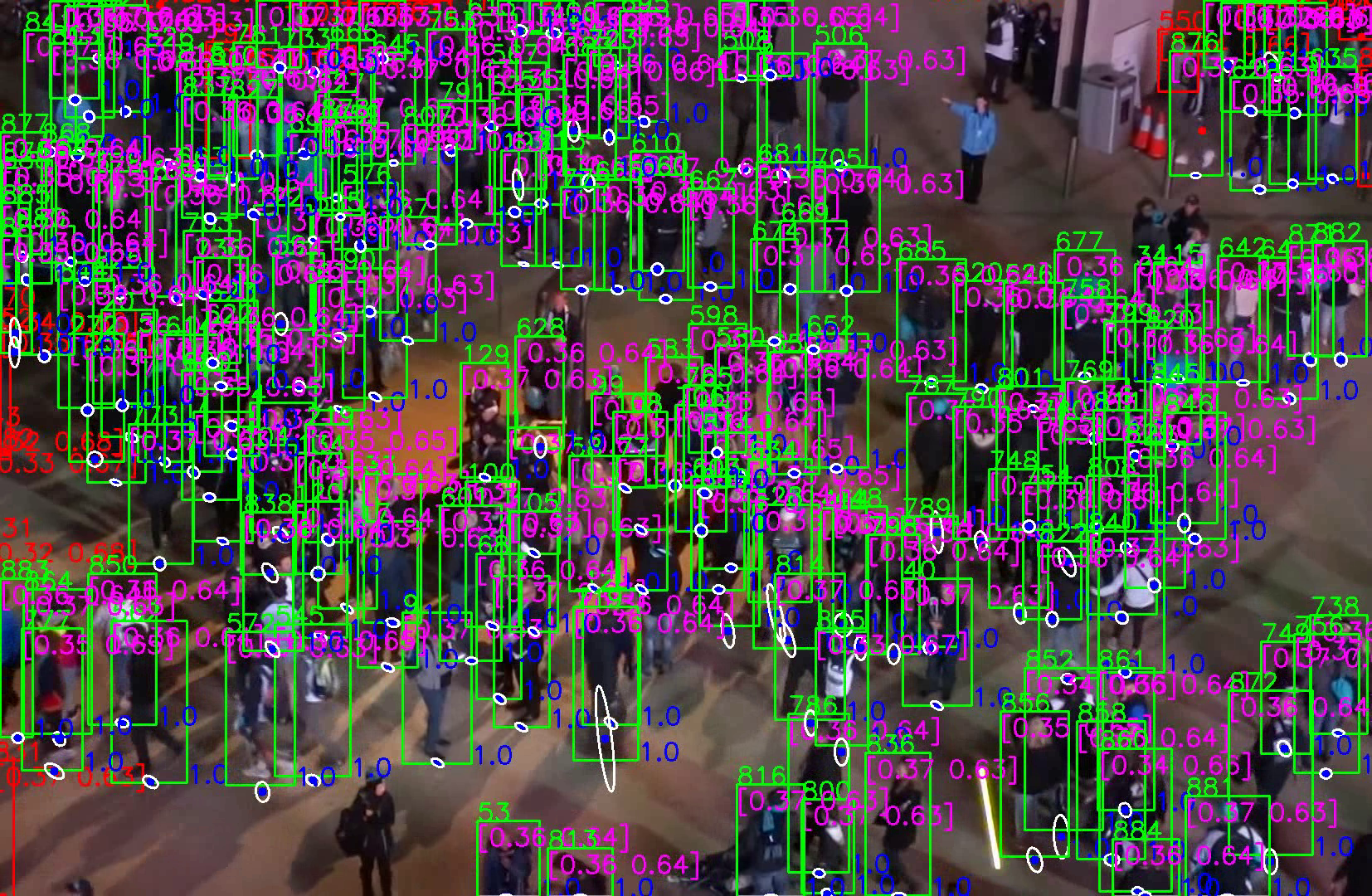}
        \caption{An example of a high track-density frame from the MOT20 dataset.}
        \label{fig:mot20crowd}
    \end{figure}

        Since MOT20 deals with even more cluttered environments than MOT17, it is unsurprising that IMM-JHSE achieves the worst results for every metric except AssA on the MOT20 test dataset in Table~\ref{tab:mot20_test}. In contrast to Figure~\ref{fig:mot17crowd}, Figure~\ref{fig:mot20crowd} shows an example of a high track-density frame from the MOT20 dataset.
        \begin{table}[h]
            \centering
            \newcolumntype{C}{>{\centering\arraybackslash} m{0.155\columnwidth} }
            \begin{tabular}{CCCCC}
            \hline
                Method & HOTA $\uparrow$ & IDF1 $\uparrow$ & MOTA $\uparrow$ & AssA $\uparrow$\\\hline

                ByteTrack \citep{Zhang2022ByteTrack:Box} & 61.3 & 75.2 & 77.8 & 59.6\\\hline
                OC-SORT \citep{Cao2022Observation-CentricTracking} & 62.4 & 76.3 & 75.7 & 62.5\\\hline
                UCMCTrack \citep{ucmc} & \textbf{62.8} & \textbf{77.4} & 75.5 & \textbf{63.5}\\\hline
                UCMCTrack+ \citep{ucmc}& \textbf{62.8} & \textbf{77.4} & \textbf{75.7} & 63.4\\\hline
                \textbf{IMM-JHSE} (Ours) & 60.87 & 74.64 & 72.82 & 61.56\\\hline
                
            \end{tabular}
            \caption{Results on the MOT20 test set with detections obtained from \citep{Zhang2022ByteTrack:Box}.}
            \label{tab:mot20_test}
        \end{table}

\section{Conclusion}\label{sec:conclusion}
    \noindent This work introduced IMM-JHSE, a novel MOT algorithm. The method uses an IMM ground plane filter, which combines static and dynamic camera motion models. Crucially, object motion is decoupled from these motion models. Furthermore, image plane information is incorporated through a simple bounding box motion filter and the BIoU score. IMM-JHSE dynamically switches between and mixes BIoU and ground plane-based Mahalanobis distance in an IMM-like fashion to perform association. Finally, dynamic process and measurement noise estimation are used. All of these components contribute to the overall performance of the method. 

    IMM-JHSE performs remarkably well compared to fully 3D MOT methods despite only using a single initial homography estimate, especially considering some 3D MOT methods on the KITTI leaderboard are offline and/or use their own detections. 

    Furthermore, IMM-JHSE performs remarkably similarly to attention-based methods on the DanceTrack dataset and outperforms them on the MOT17 dataset. Not only does IMM-JHSE bridge the gap between 3D and 2D MOT methods, but also between tracking-by-detection and tracking-by-attention methods. This could indicate that IMM-JHSE isolates some of the advantages afforded by 3D and attention-based MOT methods. Particularly, IMM-JHSE seems to indicate that the disambiguating information between tracks afforded by maintaining separate joint homography estimates (based only on an initial measurement) rivals that of 3D and attention-based methods. Crucially, IMM-JHSE does not guarantee an accurate global homography estimate, but focuses on maintaining an accurate projection for each track separately. Because IMM-JHSE requires only an initial homography estimate, the complexity of handling 3D measurements in every frame (or a context window for attention-based methods) can be foregone. However, this comes at the cost of maintaining a more complex state and association model.

    IMM-JHSE improves significantly upon related methods with the same publicly available detections, such as UCMCTrack, C-BIoU, OC-SORT and ByteTrack, on the DanceTrack and KITTI-car datasets while offering competitive performance on the MOT17, MOT20 and KITTI-pedestrian datasets. 
    
    However, a weak point of IMM-JHSE is that it performs poorly when tracks are very dense, especially in the MOT20 dataset. This is attributed to the leniency afforded by the multitude of motion models in use and the fact that IMM-JHSE is designed primarily with the DanceTrack dataset in mind. Future work may, therefore, improve its performance on data with high track densities and generalisation to other datasets. It is also possible that results may be significantly improved with the use of a custom detector module.

\section*{CRediT authorship contribution statement}
    \textbf{Paul Claasen}: Conceptualization, Methodology, Software, Validation, Formal analysis, Investigation, Resources, Data Curation, Writing - Original Draft, Visualization. \textbf{Pieter de Villiers}: Writing - Review \& Editing, Supervision, Project administration, Funding acquisition.


\section*{Data availability}
    The used datasets are already publicly available.

\section*{Acknowledgements}
    This work was supported by the MultiChoice Chair in Machine Learning and the MultiChoice Group.
    



 \bibliographystyle{elsarticle-harv} 
 \bibliography{manual_refs,general,homog,homogtracking,mht,sports}

\clearpage
\appendix
\section{IMM filter theory}\label{immappendix}
IMM-JHSE runs two ground plane EKFs in parallel, each with the state vector definition in Equation~\ref{ground_state}. One uses the static homography dynamics model in Equation~\ref{homog_static}, while the other uses the dynamic one in Equation~\ref{homog_dyn}. The IMM methodology allows the proposed method to switch between and mix state estimates from both filters, proportional to their likelihood \citep{Blom1988TheCoefficients}. This way, IMM-JHSE is robust to scenes with both static and dynamic camera movement. This section briefly describes the operation of IMM filters according to \citep{imm, imm_genovese, Blackman1999DesignSystems, Blom1988TheCoefficients}, which is implemented with the FilterPy library \citep{filterpy}. 

Consider a system with $r$ different models. The following state-space equations describe each model $i$ (assumed linear for simplicity): 
\begin{equation*}
    \begin{split}
        \mathbf{x}_t^i &= \mathbf{A}_t^i \mathbf{x}_{t-1}^i + \mathbf{B}_t^i \mathbf{u}_t^i + \mathbf{w}_t^i,\\
        \mathbf{y}_t^i &= \mathbf{G}_t^i \mathbf{x}_t^i + \mathbf{v}_t^i,
    \end{split}
\end{equation*}
where $\mathbf{x}_t^i$ is the state vector for model $i$ at time $t$, $\mathbf{A}_t^i$, $\mathbf{B}_t^i$, and $\mathbf{G}_t^i$ are the state transition, control input, and observation matrices, respectively. $\mathbf{u}_t^i$ is the control input and $\mathbf{w}_t^i$ and $\mathbf{v}_t^i$ are process and measurement noise, respectively.

The probability of each model $i$ being the correct model at time $t$ is denoted by $\mu_t^i$. These probabilities are updated at each time step based on the likelihood of the observations given each model.

The mixing probabilities $\mu_{t|t-1}^{i|j}$ represent the probability of transitioning from model $j$ at time $t-1$ to model $i$ at time $t$. These are computed using the Markov transition probabilities $p_{ji}$: 
\begin{equation*}
    \mu_{t|t-1}^{i|j} = \frac{p_{ji} \mu_{t-1}^j}{\sum_{l=1}^r p_{li} \mu_{t-1}^l}.
\end{equation*}

The mixed initial conditions for each filter are computed as follows: 
\begin{equation*}
    \begin{split}
        \hat{\mathbf{x}}_{t-1|t-1}^i &= \sum_{j=1}^r \mu_{t|t-1}^{i|j} \hat{\mathbf{x}}_{t-1|t-1}^j,\\ 
        \mathbf{P}_{t-1|t-1}^i &= \sum_{j=1}^r \mu_{t|t-1}^{i|j} \left( \mathbf{P}_{t-1|t-1}^j + (\hat{\mathbf{x}}_{t-1|t-1}^j - \hat{\mathbf{x}}_{t-1|t-1}^i)(\hat{\mathbf{x}}_{t-1|t-1}^j - \hat{\mathbf{x}}_{t-1|t-1}^i)^T \right).
    \end{split}
\end{equation*}

Each model-specific Kalman filter is updated using the mixed initial conditions:
\begin{equation*}
    \begin{split}
        \hat{\mathbf{x}}_{t|t-1}^i &= \mathbf{A}_t^i \hat{\mathbf{x}}_{t-1|t-1}^i + \mathbf{B}_t^i \mathbf{u}_t^i,\\
        \mathbf{P}_{t|t-1}^i &= \mathbf{A}_t^i \mathbf{P}_{t-1|t-1}^i (\mathbf{A}_t^i)^T + \mathbf{Q}_t^i,\\
        \mathbf{K}_t^i &= \mathbf{P}_{t|t-1}^i (\mathbf{G}_t^i)^T (\mathbf{G}_t^i \mathbf{P}_{t|t-1}^i (\mathbf{G}_t^i)^T + \mathbf{R}_t^i)^{-1},\\ 
        \hat{\mathbf{x}}_{t|t}^i &= \hat{\mathbf{x}}_{t|t-1}^i + \mathbf{K}_t^i (\mathbf{y}_t - \mathbf{G}_t^i \hat{\mathbf{x}}_{t|t-1}^i),\\ 
        \mathbf{P}_{t|t}^i &= (\mathbf{I} - \mathbf{K}_t^i \mathbf{G}_t^i) \mathbf{P}_{t|t-1}^i.
    \end{split}
\end{equation*}

The model probabilities are updated based on the likelihood of the observations: 
\begin{equation*}
    \mu_{t}^i = \frac{\Lambda_t^i \sum_{j=1}^r p_{ji} \mu_{t-1}^j}{\sum_{l=1}^r \Lambda_t^l \sum_{j=1}^r p_{jl} \mu_{t-1}^j},
\end{equation*}
where $\Lambda_t^i$ is the likelihood of the observation given model $i$.

The overall state estimate and covariance are computed as a weighted sum of the individual model estimates: 
\begin{equation*}
    \begin{split}
        \hat{\mathbf{x}}_t &= \sum_{i=1}^r \mu_t^i \hat{\mathbf{x}}_{t|t}^i,\\
        \mathbf{P}_t &= \sum_{i=1}^r \mu_t^i \left( \mathbf{P}_{t|t}^i + (\hat{\mathbf{x}}_{t|t}^i - \hat{\mathbf{x}}_t)(\hat{\mathbf{x}}_{t|t}^i - \hat{\mathbf{x}}_t)^T \right).
    \end{split}
\end{equation*}

\section{Deriving the Jacobian matrix}\label{deriv}

Let the vector $\mathbf{b}$ represent the unnormalised projection of a ground plane coordinate, i.e.
            \begin{equation*}
                \mathbf{b} = \begin{bmatrix}
                                        b_1 \\
                                        b_2 \\
                                        b_3
                                    \end{bmatrix} = \mathbf{H}^G_t\begin{bmatrix}
                                        x^{W}_t \\
                                        y^{W}_t \\
                                        1
                                    \end{bmatrix},
            \end{equation*} 
            where $x^W_t$ and $y^W_t$ represent the x- and y-coordinate of the ground plane coordinate under consideration, respectively. Expanding the above with 
            \begin{equation*}
                \mathbf{H}_t =\begin{bmatrix}
                h_{1} & h_{2} & h_{3} \\
                h_{4} & h_{5} & h_{6} \\
                h_{7} & h_{8} & h_{9} \\
            \end{bmatrix}:
            \end{equation*}
            \begin{align*}
                \begin{bmatrix}
                    b_1 \\ b_2 \\ b_3
                \end{bmatrix} = \begin{bmatrix}
                    h_1x^W_t + h_2y^{W}_t + h_3\\
                    h_4x^W_t + h_5y^{W}_t + h_6\\
                    h_7x^{W}_t + h_8y^{W}_t + h_9
                \end{bmatrix}.
            \end{align*}
The Jacobian of $\mathbf{b}$ with respect to the world coordinates is:
    \begin{equation*}
        \begin{bmatrix}
            \frac{\partial b_1}{\partial x^W_t} & \frac{\partial b_1}{\partial y^W_t}\\
            \frac{\partial b_2}{\partial x^W_t} & \frac{\partial b_2}{\partial y^W_t}\\
            \frac{\partial b_3}{\partial x^W_t} & \frac{\partial b_3}{\partial y^W_t}
        \end{bmatrix}=\begin{bmatrix}
            h_1&h_2\\h_4&h_5\\h_7&h_8
        \end{bmatrix}.
    \end{equation*}
Note that the image coordinates are obtained by $\begin{bmatrix}
    x^I_t & y^I_t
\end{bmatrix}^\top=\begin{bmatrix}
    b_1/b_3&b_2/b_3
\end{bmatrix}^\top$. Therefore, the partial derivative of $x^I_t$ with respect to $x^W_t$ is:
\begin{align*}
    \frac{\partial x^I_t}{\partial x^W_t} &= \frac{\partial b_1b_3^{-1}}{\partial x^W_t}\\
    &= \frac{h_1b_3-h_7b_1}{b_3^2}\\
    &= \gamma(h_1-h_7x^I_t),
\end{align*}
where $\gamma=b_3^{-1}$.
Similarly,
\begin{align*}
    \frac{\partial x^I_t}{\partial y^W_t}&=\gamma(h_2-h_8x^I_t),\\
    \frac{\partial y^I_t}{\partial x^W_t}&=\gamma(h_4-h_7y^I_t),\\
    \frac{\partial y^I_t}{\partial y^W_t}&=\gamma(h_5-h_8y^I_t).
\end{align*}

Thus, the Jacobian matrix of the x- and y- image coordinates with respect to the x- and y- world coordinates is
\begin{equation*}
    \mathbf{J}^{IW}_t=\begin{bmatrix}
        \frac{\partial x^I_t}{\partial x^W_t} & \frac{\partial x^I_t}{\partial y^W_t}\\
        \frac{\partial y^I_t}{\partial x^W_t} & \frac{\partial y^I_t}{\partial y^W_t}
    \end{bmatrix}=
    \gamma\begin{bmatrix}
        (h_1-h_7x^I_t) & (h_2-h_8x^I_t)\\
        (h_4-h_7y^I_t) & (h_5-h_8y^I_t)
    \end{bmatrix}.
\end{equation*}

Similarly, the Jacobian matrix of the x- and y- image coordinates with respect to the homography elements is
\begin{align*}
    \mathbf{J}^{IH}_t &= \begin{bmatrix}
        \frac{\partial x^I_t}{\partial h_1} & \frac{\partial x^I_t}{\partial h_4} & \frac{\partial x^I_t}{\partial h_7} & \frac{\partial x^I_t}{\partial h_2} & \frac{\partial x^I_t}{\partial h_5} & \frac{\partial x^I_t}{\partial h_8} & \frac{\partial x^I_t}{\partial h_3} & \frac{\partial x^I_t}{\partial h_6} & \frac{\partial x^I_t}{\partial h_9}\\
        \frac{\partial y^I_t}{\partial h_1} & \frac{\partial y^I_t}{\partial h_4} & \frac{\partial y^I_t}{\partial h_7} & \frac{\partial y^I_t}{\partial h_2} & \frac{\partial y^I_t}{\partial h_5} & \frac{\partial y^I_t}{\partial h_8} & \frac{\partial y^I_t}{\partial h_3} & \frac{\partial y^I_t}{\partial h_6} & \frac{\partial y^I_t}{\partial h_9}\\
    \end{bmatrix}\\
    &=\gamma\begin{bmatrix}
        x^W_t & 0 & -x^I_tx^W_t & y^W_t & 0 & -x^I_ty^W_t & 1 & 0 & -x^I_t\\
        0 & x^W_t & -y^I_tx^W_t & 0 & y^W_t & -y^I_ty^W_t & 0 & 1 & -y^I_t
    \end{bmatrix}.
\end{align*}

Finally, the combined Jacobian matrix of the image coordinates with respect to the homography and world coordinates is given by
\begin{equation*}
    \mathbf{J}_t = \begin{bmatrix}
        \mathbf{J}^{IW}_t & \mathbf{0}\\
        \mathbf{0} & \mathbf{J}^{IH}_t
    \end{bmatrix}.
\end{equation*}

\section{Parameter results}
Table~\ref{params} shows the parameters obtained with the pattern search algorithm for the various datasets considered. For the DanceTrack dataset, these are the parameters obtained by optimising the overall AssA. For the KITTI, MOT17 and MOT20 datasets, these are the parameters obtained by optimising the overall HOTA.
\begin{table}[h!]
\centering
\begin{tabular}{|c|c|c|c|c|}
\hline
Parameter & DanceTrack & MOT17 & MOT20 & KITTI \\
\hline
$\sigma_x$ & 5 & 5 & 0.001 & 14.84 \\
$\sigma_y$ & 18.75 & 3.6 & 5.0 & 16.25 \\
$\alpha_1$ & 0.27 & 0.91 & 0.91 & 0.83 \\
$\alpha_2$ & 0.99 & 0.46 & 0.5 & 0.5 \\
$\alpha_3$ & 0.82 & 0.99 & 0.91 & 0.91 \\
$\Omega$ & 62 & 30 & 62 & 100 \\
$b$ & 0 & 0 & 0 & 0 \\
$d_{\text{high}}$ & 0.9 & 0.6 & 0.75 & 0.29 \\
$d_{\text{low}}$ & 0.1 & 0.26 & 0.5 & 0.5 \\
$p_{s,s}$ & 0.16 & 0.99 & 0.9 & 0.72 \\
$p_{d,d}$ & 0.53 & 0.9 & 0.99 & 0.72 \\
$p_{W,W}$ & 0.9 & 0.9 & 0.72 & 0.9 \\
$p_{I,I}$ & 0.01 & 0.9 & 0.53 & 0.9 \\
$v$ & 0.56 & $1\times 10^{-7}$ & 0.005 & 0.009\\
\hline
\end{tabular}
\caption{Parameters of IMM-JHSE for different datasets.}
\label{params}

\end{table}

\end{document}